\documentclass{article}
\usepackage{lmodern}
\usepackage[T1]{fontenc}
\usepackage{iclr2027_conference,times}

\usepackage{amsmath,amsfonts}

\usepackage{graphicx}
\usepackage{booktabs}
\usepackage{hyperref}
\usepackage{url}
\usepackage{wrapfig}
\usepackage{colortbl}
\usepackage{array,tabularx,makecell,adjustbox,xcolor}
\usepackage{tikz}
\usetikzlibrary{arrows.meta,positioning,fit,calc,backgrounds,patterns,shapes.geometric}
\usepackage{needspace,placeins}
\newcounter{algorithm}
\makeatletter
\newcommand{\fnum@algorithm}{Algorithm~\thealgorithm}
\newcommand{\ext@algorithm}{loa}
\makeatother
\newcolumntype{Y}{>{\raggedright\arraybackslash}X}

\definecolor{accentblue}{HTML}{24658B}
\definecolor{accentteal}{HTML}{187563}
\definecolor{accentpurple}{HTML}{71529C}
\definecolor{ink}{HTML}{22364A}
\definecolor{softblue}{HTML}{E9F0F5}
\definecolor{softpurple}{HTML}{EEEAF5}
\definecolor{softteal}{HTML}{E8F3F0}
\definecolor{softorange}{HTML}{FBF0E4}
\definecolor{rulegray}{HTML}{9AA8B3}
\definecolor{notegray}{HTML}{52616B}
\definecolor{bestgreen}{HTML}{EAF7EE}    % LayerEditor row
\definecolor{secondblue}{HTML}{EEF5FC}   % second best
\definecolor{rulegreen}{HTML}{B8DDC2}    % separator above LayerEditor
\tikzset{
 block/.style={draw=rulegray,rounded corners=3pt,line width=0.55pt,align=center,inner sep=5pt,font=\scriptsize},
 arrow/.style={-{Latex[length=2mm,width=1.3mm]},line width=0.7pt,draw=ink},
 pbox/.style={draw=rulegray,dashed,rounded corners=2pt,fill=black!2,align=center,font=\scriptsize,inner sep=3pt}
}
\hypersetup{hidelinks,pdftitle={Editing Images into Reusable Layers via Compositional Edit-Prior Distillation},pdfauthor={Wen Xue, Yong Liu, Xueheng Li, Si Wu, Yong Xu, Yipeng Sun, Yan Li, Simiu Gu}}
\ifdefined\pdfminorversion\pdfminorversion=7\fi
\ifdefined\XeTeXversion\AtBeginDocument{}\fi

\iclrfinalcopy
\ifdefined\XeTeXversion
  \catcode"021B=13
  \defț{\c{t}}
\fi

\title{Editing Images into Reusable Layers\\via Compositional Edit-Prior Distillation}
\author{%
  \parbox[t]{\dimexpr\textwidth-2\tabcolsep\relax}{\centering\normalfont
  \textbf{Wen Xue}\textsuperscript{1,*}\quad
  \textbf{Yong Liu}\textsuperscript{2,*}\quad
  \textbf{Xueheng Li}\textsuperscript{3}\quad
  \textbf{Si Wu}\textsuperscript{1}\\[3pt]
  \textbf{Yong Xu}\textsuperscript{4}\quad
  \textbf{Yipeng Sun}\textsuperscript{5,\textdagger}\quad
  \textbf{Yan Li}\textsuperscript{5}\quad
  \textbf{Simiu Gu}\textsuperscript{5}\\[7pt]
  {\normalfont\small\textsuperscript{1}South China University of Technology\quad
  \textsuperscript{2}Tsinghua University}\\[2pt]
  {\normalfont\small\textsuperscript{3}University of Science and Technology of China}\\[2pt]
  {\normalfont\small\textsuperscript{4}Peng Cheng Laboratory\quad
  \textsuperscript{5}JD.com}
}}

\date{}

\begin{document}
\raggedbottom
\maketitle
\lhead{Preprint}
\begingroup
\renewcommand{\thefootnote}{}
\makeatletter
\long\def\@makefntext#1{\noindent#1}
\makeatother
\begin{NoHyper}
\footnotetext{\fontsize{8}{9}\selectfont\textsuperscript{*}Equal contribution.\quad\textsuperscript{\textdagger}Corresponding author.}
\end{NoHyper}
\endgroup

\begin{abstract}
High-quality image editing requires precise modification of target regions while preserving the content of unrelated regions. 
Existing editing models produce flattened outputs that limit precise local manipulation and layer reuse, while decomposition methods typically optimize layer recovery separately from instruction-driven editing.
To this end, we introduce \textbf{LayerEditor}, a unified framework that couples instruction-guided image editing with controllable generation of reusable RGBA layers. At its core, \textbf{In-Context RGB--Alpha Token Modeling} jointly represents the edited canvas and an ordered sequence of RGB--Alpha layers within a single token sequence, enabling semantic edits and explicit control over individual layer content. To retain the pretrained model's editing capability during adaptation to layered outputs, we propose \textbf{Compositional Edit-Prior Distillation}, which transfers its editing prior to both the edited canvas and the image composed from the predicted layers. To support training and evaluation, we construct \textbf{LayerEditor-100K}, a large-scale paired image-editing dataset with ordered RGBA layer annotations, and introduce dedicated benchmarks for layered editing. Extensive experiments demonstrate that LayerEditor maintains strong editing fidelity while generating accurate RGBA layers, supporting both flexible layer-level manipulation and multi-layer decomposition.
\end{abstract}

\section{Introduction}
\label{sec:introduction}

Instruction-driven image editing aims to modify specified content while preserving unrelated regions of the source image. Recent editing models~\citep{wu2025qwenimage,firered2026rededit} commonly employ multimodal diffusion transformers to jointly model textual instructions and images within a unified token sequence, synthesizing a single flattened edited canvas in which text, objects, and background content are fused together. Such flattened outputs make it difficult to manipulate fine-grained details or incorporate detailed user-provided elements, which is crucial for downstream applications such as e-commerce content production and advertising design. Subsequent revisions therefore require repeated generation, increasing processing overhead and risking unintended changes to unrelated content and boundary artifacts.

Image decomposition provides a natural way to recover explicit layers from such flattened results~\citep{yin2025qwenlayered,he2026liwi,tang2026mrt}. Decomposition and image editing can be combined into layered workflows, as illustrated by Agent Banana~\citep{ye2026agentbanana}. Yet editing and decomposition are typically optimized as separate stages with different objectives. Editing models focus on producing visually faithful results without considering the desired layer structure, while decomposition models recover layers after editing without jointly optimizing the requested edit and target layer structure. Consequently, even visually plausible edits may yield incomplete or poorly organized layers that are difficult to reuse. Correcting these layer-level errors often requires repeated model calls and manual refinement, increasing labor costs while reducing the efficiency and scalability of layered editing. 
These limitations motivate a unified formulation that jointly performs instruction-driven editing and reusable layer generation.

\begin{figure}[tbp]
 \centering
 \includegraphics[width=\linewidth]{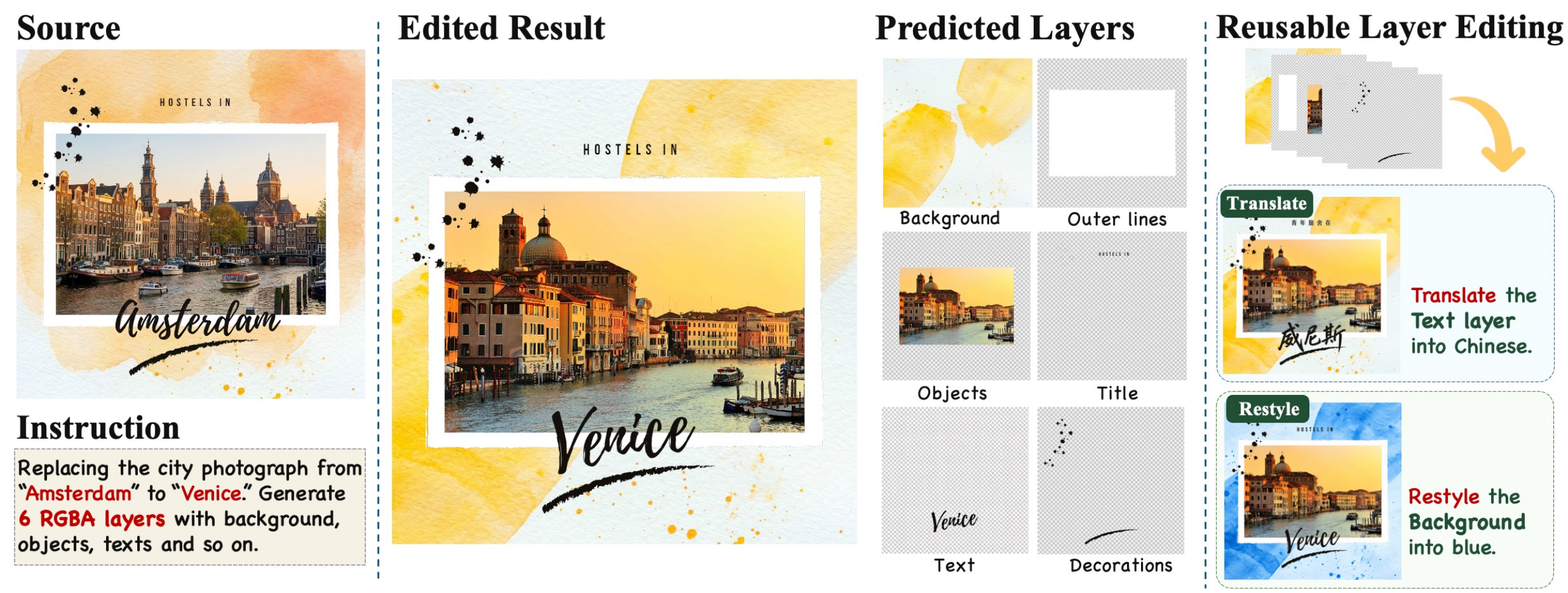}
 
 \caption{LayerEditor edits images into reusable RGBA layers with user-specified layer content. The predicted layers can be independently manipulated for subsequent edits.}
 \label{fig:teaser}
\end{figure}

In this work, we introduce LayerEditor, a unified model that couples instruction-driven image editing with controllable generation of reusable RGBA layers, as illustrated in Figure~\ref{fig:teaser}.
We propose In-Context RGB--Alpha Token Modeling to jointly represent the edited canvas and ordered RGB--Alpha layers within one token sequence. A dedicated Alpha decoder further recovers continuous opacity for fine structures and semitransparent boundaries. Moreover, directly extending a pretrained editing model to layered outputs tends to degrade its original editing capability, i.e., the edited content exhibits quality drops or artifacts. To mitigate this degradation, we introduce on-policy Compositional Edit-Prior Distillation (C-EPD) which facilitates the model's adaptation to layered outputs while maintaining its pre-trained generative editing capacity. C-EPD queries a frozen editing teacher at intermediate states along the student's denoising trajectory, where the teacher supervises the direct edited canvas and the composition of layers predicted by the layered editing student. Extensive experiments demonstrate that LayerEditor preserves strong editing performance while supporting flexible layered outputs with variable layer counts and user-specified layer content.

To further provide paired supervision for both editing outcomes and their underlying layer structures, we construct \textbf{LayerEditor-100K}, a large-scale dataset tailored to instruction-driven layered image editing. Each example contains a source image, editing instruction, edited target, and ordered RGBA layer stack, covering diverse text, object, and appearance edits with variable layer counts. In addition, since existing benchmarks do not adequately evaluate instruction-driven layered editing, we construct \textbf{\mbox{Crello-Testset}} and \textbf{\mbox{ScaleEdit-Testset}}, with dedicated settings for layer-count control and user-specified layer content, enabling systematic evaluation of editing fidelity, layer quality, and layered controllability.

\begin{itemize}
    \item We introduce LayerEditor, a unified model for editing images into reusable RGBA layers through In-Context RGB--Alpha Token Modeling. LayerEditor supports variable layer counts and user-specified layer content.

    \item We construct LayerEditor-100K, a large-scale dataset tailored to layered editing, containing annotated image pairs with ordered RGBA layers, together with two benchmarks, \mbox{Crello-Testset} and \mbox{ScaleEdit-Testset}, for evaluating layered editing quality and controllability.

    \item We propose C-EPD, an on-policy distillation strategy that transfers the pretrained editing prior to layered outputs, preserving editing fidelity during layered adaptation.
\end{itemize}
\section{Related Work}
\label{sec:related_work}

\subsection{Conditional Image Editing}

Image editing aims to modify a source image according to textual instructions while preserving unrelated content. InstructPix2Pix~\citep{brooks2023instructpix2pix} established paired image--instruction supervision for instruction-driven editing, while Qwen-Image~\citep{wu2025qwenimage} and FireRed-Image-Edit~\citep{firered2026rededit} further improved editing fidelity and instruction following with stronger generative priors. Related methods explored selective reconstruction, interactive control, and temporal consistency for more fine-grained manipulation~\citep{wen2024retouchformer,xue2025retouchgpt,xue2024vretoucher}. More recently, agentic systems extended image editing into multi-step planning and execution workflows. Visual ChatGPT~\citep{wu2023visualchatgpt} coordinated specialized visual tools through language-based planning, PhotoAgent~\citep{yao2026photoagent} performed iterative aesthetic editing with search and visual feedback, and Agent Banana~\citep{ye2026agentbanana} incorporated hierarchical planning and layer decomposition for localized editing. Despite improved editing quality and control, these approaches either produce flattened outputs or rely on cascaded workflows to recover editable layers. 

\subsection{Layered Image Generation}
Existing layered visual modeling mainly focuses on layered generation and structured editing. LayerDiff~\citep{huang2024layerdiff}, PSDiffusion~\citep{huang2025psdiffusion}, LayeringDiff~\citep{kang2025layeringdiff}, and LASAGNA~\citep{yang2026lasagna} generate coordinated layers from text or spatial conditions. For graphic design, COLE~\citep{jia2023cole}, CreatiPoster~\citep{zhang2025creatiposter}, and Editable Visual Design~\citep{ye2026editablevisualdesign} organize generated assets into editable structures, while MiLDEdit~\citep{lin2026mildedit} and MRT~\citep{tang2026mrt} extend generation and editing to layered inputs. More recently, LaDe~\citep{lungu2026lade} jointly generates design canvases and RGBA layers, and UniWorld-Design~\citep{li2026uniworld} supports layered generation and prompt-controlled decomposition. These methods emphasize layered generation and decomposition rather than jointly editing a flattened source image and generating its edited layer stack. Image decomposition methods such as Qwen-Image-Layered~\citep{yin2025qwenlayered} and LiWi~\citep{he2026liwi} recover editable components from rendered images, with related transparency-aware methods improving opacity and boundary quality~\citep{rhemann2009alphamatting,zhang2024layerdiffuse,yu2026omnialpha}. Decomposition is typically optimized separately from instruction-driven editing. LayerEditor addresses this gap by jointly editing a flattened source image and generating the corresponding reusable RGBA layers.

\subsection{On-Policy Distillation}
On-policy distillation (OPD) transfers knowledge from a teacher to a student by querying the teacher online on states actually visited by the student. By supervising the student's own trajectories, OPD reduces the mismatch between teacher-generated training data and the student's evolving state distribution~\citep{agarwal2024gkd}. Recent studies extended this idea to flow-based generative models. Flow-OPD~\citep{fang2026flowopd} distilled specialized teachers through student rollouts and manifold anchoring, DiffusionOPD~\citep{li2026diffusionopd} studied continuous-state transition alignment, and STEP-OPD~\citep{wei2026stepopd} further explored extrapolated targets and intermediate representation alignment. Unlike progressive distillation~\citep{salimans2022progressive} and consistency models~\citep{song2023consistency}, which mainly reduced the number of sampling steps, our goal is to preserve editing capability when extending a pretrained editor from single-image outputs to structured layered outputs. 

\section{Method}
\label{sec:method}
\label{sec:task}
\label{sec:architecture}

\begin{figure}[tbp]
    \centering
    \includegraphics[width=\linewidth]{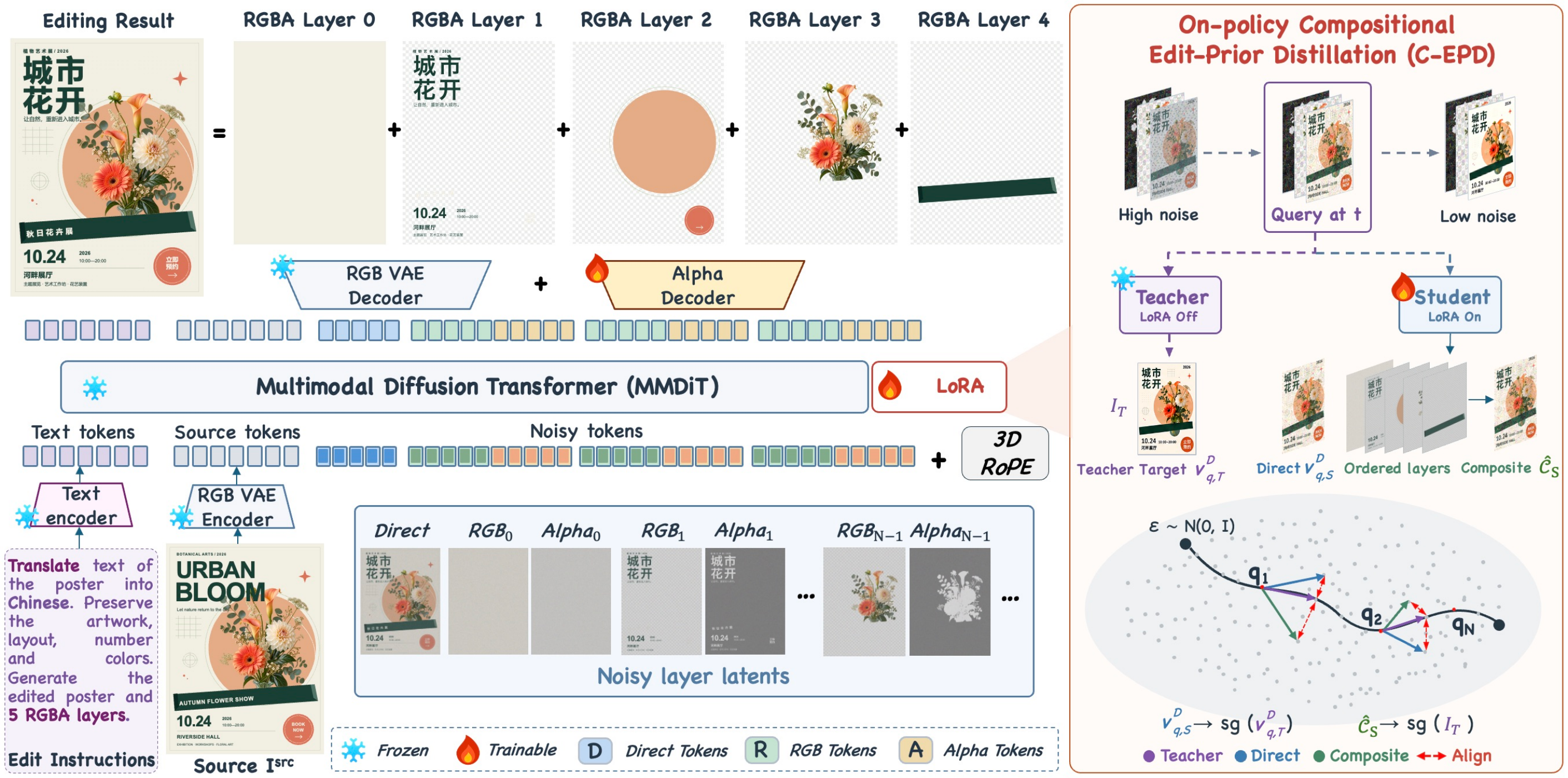}
    
    \caption{Overview of LayerEditor.
    LayerEditor jointly generates the edited canvas $\mathcal D$ and ordered RGBA stack $\mathcal S$. C-EPD uses the frozen teacher to supervise both the Direct prediction and the decoded composition reconstructed from the predicted layers. }
    \label{fig:flowchart}
\end{figure}

Given a source image $I^{\mathrm{src}}$, an editing instruction $u$, and a requested layer count $N$, LayerEditor jointly generates an edited canvas and $N$ reusable RGBA layers, as illustrated in Figure~\ref{fig:flowchart}.
\begin{equation}
    (\mathcal D,\mathcal S)
    =
    G(I^{\mathrm{src}},u,N),
    \qquad
    \mathcal S
    =
    \{\mathcal L_i=(R_i,A_i)\}_{i=0}^{N-1}.
    \label{eq:task}
\end{equation}
Here, $\mathcal D$ denotes the directly generated edited canvas, $G$ denotes the generative model, $R_i$ and $A_i$ represent the RGB appearance and Alpha opacity of layer $i$. The layer stack $\mathcal S$ is ordered from back to front, and its exact alpha-over rendering produces the composited image $\mathcal C=\operatorname{Comp}(\mathcal S)$. The editing instruction $u$ may additionally specify desired content for individual layers.

\subsection{In-Context RGB--Alpha Token Modeling}
\label{sec:joint_representation}

LayerEditor extends a pretrained image editing model~\citep{firered2026rededit} from single-image generation to joint image-and-layer generation. Frozen encoders provide source-image and instruction conditions, while the requested layer count $N$ determines the number of RGB--Alpha output pairs. At noise level $\sigma_t$, the joint latent state is
\begin{equation}
    Z_t=
    [
    z_t^{\mathcal D},
    z_t^{R_0},z_t^{A_0},
    \ldots,
    z_t^{R_{N-1}},z_t^{A_{N-1}}
    ].
    \label{eq:joint_state}
\end{equation}
All output fields are initialized from Gaussian noise and jointly denoised within a unified sequence. Building on sequence-based layered generation~\citep{yin2025qwenlayered}, In-Context RGB--Alpha Token Modeling represents appearance and opacity as distinct interacting fields. Shared attention captures dependencies across the edited canvas $\mathcal D$ and layer stack $\mathcal S$. We adopt layer-aware rotary coordinates $(s,x,y)$ to distinguish fields sharing the same canvas coordinates. Source-image tokens use $s=-1$, the edited canvas uses $s=0$, and the RGB and Alpha fields of layer $i$ use $s=2i+1$ and $s=2i+2$, respectively. The slot coordinate encodes field identity and stacking order, while $(x,y)$ preserves spatial correspondence. This formulation supports variable layer counts within a shared architecture.
\label{sec:alpha_decoder}
We further introduce a dedicated Alpha decoder to recover continuous opacity $A_i$. It is initialized from the pretrained RGB decoder and trained separately with opacity and compositional reconstruction objectives. Separating the RGB and Alpha decoders preserves the pretrained RGB pathway while specializing transparency reconstruction for fine structures and semitransparent boundaries. Details of Alpha decoder training are provided in Appendix~\ref{app:latent_decoder}.

\subsection{Compositional Edit-Prior Distillation}
\label{sec:opd}
Adapting a pretrained image editing model to layered outputs tends to weaken its original editing capability. As shown by the w/o C-EPD ablative results in Figure~\ref{fig:ablation_qualitative}, layered SFT alone may introduce text corruption, structural distortions, and unintended changes to surrounding content. We therefore propose C-EPD, which uses the frozen pretrained editing model as an on-policy teacher to preserve the original editing prior. The teacher guides both the Direct output and the composed layered representation, mitigating editing degradation during layered adaptation.

Following DiffusionOPD~\citep{li2026diffusionopd}, C-EPD queries the frozen teacher at intermediate states from a detached student denoising trajectory. At a query state $\bar Z_q$ with noise level $\sigma_q$, the student recomputes Direct, RGB, and Alpha velocities with gradients. The teacher receives the same Direct latent $\bar z_q^{\mathcal D}$, source image, and editing instruction to predict $v_{q,T}^{\mathcal D}$. This supplies on-policy guidance at states visited by the adapted model. For each student output field $X\in\{R_i,A_i\}$, we estimate its local clean latent from the current noisy state and predicted velocity as $\hat z_{0|q}^{X}=\bar z_q^{X}-\sigma_qv_{q,S}^{X}$. Frozen decoders recover $\hat R_i=\mathcal D_R(\hat z_{0|q}^{R_i})$ and $\hat A_i=\mathcal D_A(\hat z_{0|q}^{A_i})$. Starting from a white background $\widehat{\mathcal C}_{-1}=B$, we compose these predictions in back-to-front order:
\begin{equation}
\widehat{\mathcal C}_i=\hat A_i\odot\hat R_i+(1-\hat A_i)\odot\widehat{\mathcal C}_{i-1},\qquad \widehat{\mathcal C}_S=\widehat{\mathcal C}_{N-1}.
\label{eq:pixel_composite}
\end{equation}
Here the student's predicted Alpha controls each layer's visible contribution through standard source-over composition. The approximation lies in the local clean estimates. At the same query $q$, $\widehat{\mathcal C}_S$ denotes the student composite and $\mathcal I_T=\operatorname{sg}[\mathcal D_R(\bar z_q^{\mathcal D}-\sigma_qv_{q,T}^{\mathcal D})]$ is the decoded teacher target. We retain Direct velocity matching and supervise the decoded composite:
\begin{equation}
\mathcal L_{\mathrm{C\text{-}EPD}}
=
\mathbb E_{q\sim\mathcal Q}
\left[
\frac{1}{d_{\mathcal D}}
\left\|
v_{q,S}^{\mathcal D}
-
\operatorname{sg}(v_{q,T}^{\mathcal D})
\right\|_2^2
+
\frac{\lambda_{\mathcal S}}{d_{\mathcal C}}
\left\|
\widehat{\mathcal C}_S-\mathcal I_T
\right\|_2^2
\right].
\label{eq:composite_opd}
\end{equation}
Here $d_{\mathcal D}$ and $d_{\mathcal C}$ count the elements of the Direct velocity field and decoded composite, respectively, and $\operatorname{sg}$ stops gradients through the teacher target. The first term aligns the student's Direct velocity with the frozen teacher, while the second transfers the same editing prior to the layered output through its decoded composition. $\lambda_{\mathcal S}$ balances the compositional supervision. The gradients from the second term are propagated through student-side decoding and alpha compositing to the layered prediction branch.

\begin{figure}[tbp]
    \centering
    \includegraphics[width=\linewidth]{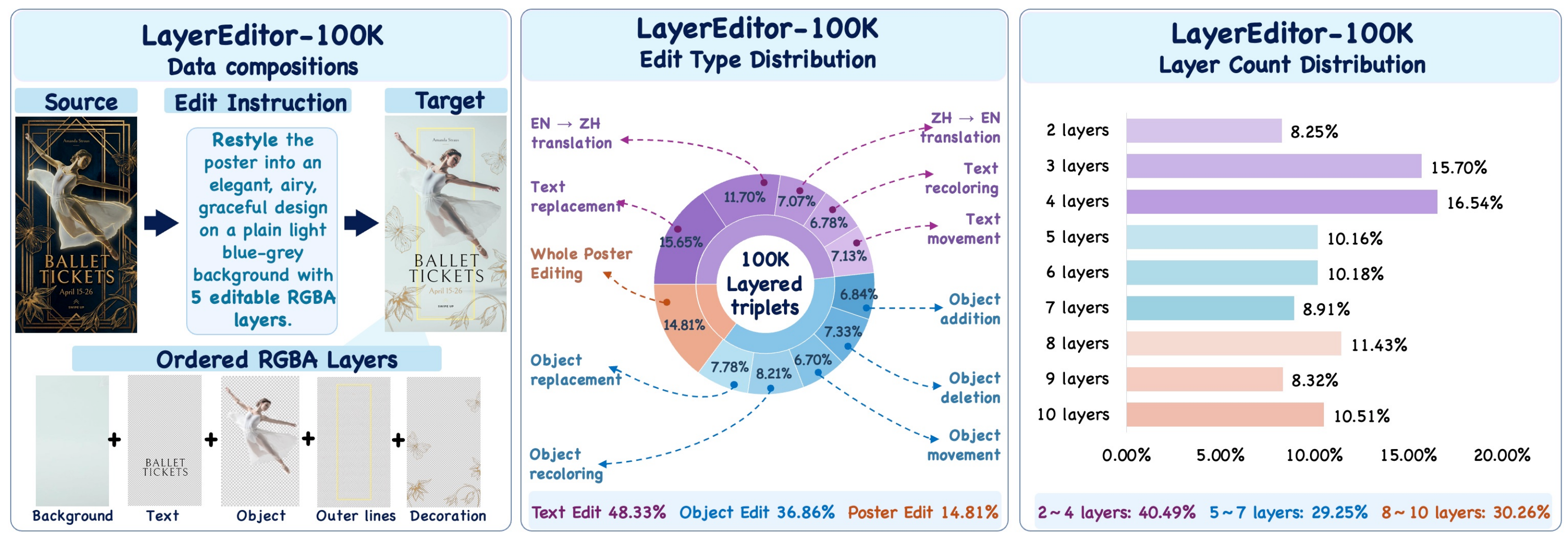}
    \caption{Overview of LayerEditor-100K. Paired editing examples and ordered RGBA annotations cover text, object, and poster-level appearance edits with $2$--$10$ layers, providing supervision across diverse editing tasks and layer structures.}
    \label{fig:data_distribution}
\end{figure}

\subsection{Training Objectives}
\label{sec:objectives}

LayerEditor adapts the frozen pretrained MMDiT with LoRA~\citep{hu2021lora}. Given the clean target sequence
$Z^*=[z^{\mathcal D,*},z^{R_0,*},z^{A_0,*},\ldots,z^{R_{N-1},*},z^{A_{N-1},*}]$,
we construct the joint noisy state
$Z_t=(1-\sigma_t)Z^*+\sigma_t\epsilon$
with target velocity $v_t^*=\epsilon-Z^*$.
The model predicts the joint velocity $v_\theta$ from $Z_t$ conditioned on the source image, editing instruction, requested layer count, and noise level. We use the flow-matching loss to supervise the complete velocity sequence: 
\begin{equation}
    \mathcal L_{\mathrm{flow}}
    =
    \mathbb E_{t,\epsilon}
    \left[
        \frac{1}{d_Z}
        \left\|
            v_\theta(Z_t)
            -
            v_t^*
        \right\|_2^2
    \right],
    \label{eq:flow}
\end{equation}
where $d_Z$ denotes the number of elements in the joint output sequence. After the initial flow-matching warm-up, C-EPD introduces teacher guidance for the Direct velocity and decoded layer composition while flow matching remains active. The combined objective is as follows, where $\lambda_{\mathrm C}$ controls the strength of editing-prior guidance: 
\begin{equation}
    \mathcal L_\theta
    =
    \mathcal L_{\mathrm{flow}}
    +
    \lambda_{\mathrm C}\mathcal L_{\mathrm{C\text{-}EPD}}.
    \label{eq:lora_objective}
\end{equation}

\subsection{LayerEditor-100K and Evaluation Benchmarks}
\label{sec:dataset_benchmark}
\label{sec:dataset_construction}

LayerEditor-100K contains 100K editing pairs constructed from Crello~\citep{yamaguchi2021canvasvae}, PrismLayersPro~\citep{chen2025prismlayers}, and ScaleEdit~\citep{chen2026scaleedit}. Each example provides a source image, editing instruction, edited target, and ordered full-canvas RGBA stack that reconstructs the target. Native Crello and PrismLayersPro designs retain their layered structure during edit construction. Flattened ScaleEdit pairs acquire target layers through planning, decomposition, refinement, and quality filtering. Both routes verify instruction execution, layer isolation, and recomposition.

Figure~\ref{fig:data_distribution} summarizes 48,329 text edits, 36,858 object edits, and 14,813 poster-level appearance edits. Text operations include movement, recoloring, replacement, and translation. Object operations cover addition, removal, movement, recoloring, and replacement. Each target contains $2$--$10$ layers, spanning compact layouts and fine-grained organizations of text, objects, and decorations.

To evaluate editing fidelity and layer usability, we construct two disjoint benchmarks, \textbf{\mbox{Crello-Testset}} with 2,994 cases derived from native Crello and PrismLayersPro designs and \textbf{\mbox{ScaleEdit-Testset}} with 1,000 cases derived from flattened editing pairs with recovered target layers. Both benchmarks are constructed independently from the training data, with source documents and all derived editing or layer variants kept strictly disjoint. We further pool the 3,994 evaluation cases to construct controlled settings that assess complementary aspects of layered controllability. The layer-count control setting varies the requested layer count and reference organization while preserving the source image and intended edited appearance, testing whether the model can reorganize the same content at different layer granularities. In addition, we evaluate image-to-layer decomposition (I2L) on held-out original Crello examples with at most ten layers to assess layer recovery ability.

\section{Experiments}
\label{sec:experimental_setup}
\setcounter{topnumber}{2}
\setcounter{totalnumber}{3}
\setlength{\textfloatsep}{8pt plus 2pt minus 2pt}
\setlength{\intextsep}{6pt plus 2pt minus 2pt}

\subsection{Experiment Setups}
\paragraph{Implementation Details}
LayerEditor is initialized from FireRed-Image-Edit-1.1 and adapted using rank-256 LoRA, with a dedicated Alpha decoder for opacity reconstruction. Training begins with two epochs of $\mathcal L_{\mathrm{flow}}$, followed by joint optimization with C-EPD using AdamW at a learning rate of $2\times10^{-5}$ with a batch size of 64. Images retain their aspect ratios within a $1024\times1024$ pixel budget. Further details are provided in Appendix~\ref{app:implementation}.

\begin{table}[tbp]
\centering
\caption{Editing and layer quality across two benchmarks. Rubric scores assess Composite outputs (1--10). The \protect\colorbox{bestgreen}{\textbf{best}} and \protect\colorbox{secondblue}{\underline{second-best}} reported values are highlighted in each column.}
\label{tab:main}

\begingroup
\fontsize{7.3}{8.1}\selectfont
\setlength{\tabcolsep}{2.2pt}
\setlength{\fboxsep}{1.2pt}
\renewcommand{\arraystretch}{1.04}

\newcommand{\best}[1]{\cellcolor{bestgreen}\textbf{#1}}
\newcommand{\secondbest}[1]{\cellcolor{secondblue}#1}

\begin{adjustbox}{max width=\linewidth}
\begin{tabular}{@{}l*{11}{c}@{}}
\toprule
&\multicolumn{4}{c}{\textcolor{accentblue}{\textbf{Image level}}}
&\multicolumn{2}{c}{\textcolor{accentteal}{\textbf{Layer level}}}
&\multicolumn{5}{c}{\textcolor{accentpurple}{\textbf{Rubric-based scores $\uparrow$ (1--10)}}}\\
\cmidrule(lr){2-5}\cmidrule(lr){6-7}\cmidrule(l){8-12}

Method
&\makecell{Direct\\SSIM$\uparrow$}
&\makecell{Direct\\FID$\downarrow$}
&\makecell{Comp.\\SSIM$\uparrow$}
&\makecell{Comp.\\FID$\downarrow$}
&\makecell{RGBA\\L1$\downarrow$}
&\makecell{Alpha\\IoU$\uparrow$}
&\makecell{Edit\\Consist.}
&\makecell{Visual\\Quality}
&\makecell{Layer\\Segment.}
&\makecell{Layer\\Compose.}
&\makecell{Alpha\\Accuracy}\\
\midrule

\multicolumn{12}{@{}l}{\textcolor{accentblue}{\textbf{\mbox{Crello-Testset} (2,994 documents)}}}\\
Seedream5-Pro (Seq.)&\cellcolor{bestgreen}\textbf{0.8586}&\cellcolor{secondblue}\underline{33.04}&0.7542&61.89&0.1366&0.4980&4.46&4.75&5.13&3.87&{3.84}\\
Seedream5-Pro (Joint)&--&--&\cellcolor{secondblue}\underline{0.7902}&\cellcolor{secondblue}\underline{55.33}&0.1344&\cellcolor{secondblue}\underline{0.5139}&4.14&5.10&\cellcolor{secondblue}\underline{5.22}&4.17&\cellcolor{secondblue}\underline{3.85}\\
FR + QIL&0.6491&46.62&0.5584&115.99&0.1417&0.2208&1.48&1.73&4.03&2.74&1.78\\
FR + QIL-Control&0.6491&46.62&0.5522&120.20&0.2447&0.1246&1.67&2.42&3.02&1.75&1.17\\
FR + QIL\textit{-Decomp-FT}&0.6491&46.62&0.6343&57.15&0.2398&0.2047&\cellcolor{secondblue}\underline{5.12}&\cellcolor{secondblue}\underline{6.01}&4.91&\cellcolor{secondblue}\underline{4.35}&1.22\\
QIL\textit{-Edit-SFT}&0.7404&56.85&0.7285&65.55&\cellcolor{secondblue}\underline{0.0993}&0.3171&3.55&4.34&4.55&3.89&1.59\\
RevealLayer&0.6491&46.62&0.6165&67.55&0.5392&0.1900&4.26&4.19&3.38&3.12&1.18\\
\midrule
\rowcolor{bestgreen}\textbf{LayerEditor}&\cellcolor{secondblue}\underline{0.8530}&\cellcolor{bestgreen}\textbf{30.15}&\cellcolor{bestgreen}\textbf{0.8351}&\cellcolor{bestgreen}\textbf{33.86}&\cellcolor{bestgreen}\textbf{0.0350}&\cellcolor{bestgreen}\textbf{0.6074}&\cellcolor{bestgreen}\textbf{6.14}&\cellcolor{bestgreen}\textbf{6.35}&\cellcolor{bestgreen}\textbf{6.83}&\cellcolor{bestgreen}\textbf{5.64}&\cellcolor{bestgreen}\textbf{5.05}\\
\midrule
\multicolumn{12}{@{}l}{\textcolor{accentpurple}{\textbf{\mbox{ScaleEdit-Testset} (1,000 documents)}}}\\
Seedream5-Pro (Seq.)&\cellcolor{bestgreen}\textbf{0.8933}&\cellcolor{bestgreen}\textbf{35.53}&0.7901&70.69&{0.0585}&0.6327&4.65&4.80&5.33&4.05&\cellcolor{secondblue}\underline{4.60}\\
Seedream5-Pro (Joint)&--&--&\cellcolor{secondblue}\underline{0.8365}&59.62&\cellcolor{secondblue}\underline{0.0552}&\cellcolor{secondblue}\underline{0.6583}&4.05&4.71&\cellcolor{secondblue}\underline{5.49}&4.13&4.55\\
FR + QIL&0.7088&41.62&0.6451&81.09&0.3009&0.2044&4.92&4.76&4.23&3.81&1.44\\
FR + QIL-Control&0.7088&41.62&0.6606&84.07&0.1791&0.4899&4.70&5.08&5.25&4.06&2.20\\
FR + QIL\textit{-Decomp-FT}&0.7088&41.62&0.6927&\cellcolor{secondblue}\underline{49.37}&0.1711&0.4240&\cellcolor{secondblue}\underline{5.40}&\cellcolor{secondblue}\underline{5.29}&5.01&\cellcolor{secondblue}\underline{5.16}&1.42\\
QIL\textit{-Edit-SFT}&0.7972&70.01&0.7895&77.74&0.1982&0.3261&3.86&4.41&4.32&4.12&1.32\\
RevealLayer&0.7088&41.62&0.6464&75.74&0.3070&0.4465&4.28&4.08&3.56&3.07&1.10\\
\midrule
\rowcolor{bestgreen}\textbf{LayerEditor}&\cellcolor{secondblue}\underline{0.8925}&\cellcolor{secondblue}\underline{36.16}&\cellcolor{bestgreen}\textbf{0.8839}&\cellcolor{bestgreen}\textbf{39.48}&\cellcolor{bestgreen}\textbf{0.0515}&\cellcolor{bestgreen}\textbf{0.6802}&\cellcolor{bestgreen}\textbf{5.96}&\cellcolor{bestgreen}\textbf{6.58}&\cellcolor{bestgreen}\textbf{6.26}&\cellcolor{bestgreen}\textbf{5.84}&\cellcolor{bestgreen}\textbf{5.13}\\
\bottomrule
\end{tabular}
\end{adjustbox}
\endgroup
\par\smallskip
\begin{minipage}{\linewidth}
\scriptsize\raggedright
\textit{Note.} FR: FireRed-Image-Edit-1.1. QIL: Qwen-Image-Layered. Due to platform restrictions, Seedream5-Pro failed to produce usable outputs for five \mbox{Crello-Testset} documents in both the Seq. and Joint settings. The corresponding metrics exclude these missing cases and are computed on the remaining documents. Both settings of Seedream5-Pro use the official API.
\end{minipage}
\end{table}

\paragraph{Competing Methods}
LayerEditor is compared with representative image editing and layer-decomposition baselines using their default or official inference settings. QIL, QIL-Decomp-FT, and RevealLayer use the requested layer count without content descriptions. QIL-Control, QIL-Edit-SFT, Seedream5-Pro, and LayerEditor additionally receive the same ordered layer-content specifications. Seedream5-Pro~\citep{bytedance2026seedream5pro} is evaluated through its official API under two settings: a sequential pipeline that performs editing followed by decomposition, and a joint setting that requests editing and layer decomposition in a single API call. We also construct cascaded baselines by combining FireRed-Image-Edit-1.1 (FR)~\citep{firered2026rededit} with Qwen-Image-Layered (QIL)~\citep{yin2025qwenlayered}, QIL-Control~\citep{diffsynth2026layeredcontrol}, or RevealLayer~\citep{wang2026reveallayer}. QIL-Decomp-FT follows the official QIL training pipeline on LayerEditor-100K, while QIL-Edit-SFT directly learns editing and layered generation. Detailed inference settings are in Appendix~\ref{app:baselines}.

\paragraph{Evaluation Protocols}
We evaluate layered editing performance primarily through the recomposed image $\mathcal C$, since the requested edit should remain correct after the predicted RGBA layers are combined. Direct and Composite SSIM/FID~\citep{wang2004ssim,heusel2017fid} measure similarity to the edited target, with the Direct output serving as an auxiliary canvas diagnostic. RGBA L1 and Alpha IoU assess reference-layer agreement after matching and ordered pairing of remaining layers, as detailed in Appendix~\ref{app:metrics}. To complement reference-based metrics, we use a unified five-dimensional rubric with VLM-assisted scoring for all methods. The rubric assesses edit consistency, visual quality, layer segmentation, layer composition, and Alpha accuracy from the recomposed output together with the corresponding target and reference layers. Figure~\ref{fig:rubric_instructions} and Appendix~\ref{app:evaluation} provide the detailed scoring protocol.

\begin{figure}[tbp]
\centering
\includegraphics[width=\linewidth]{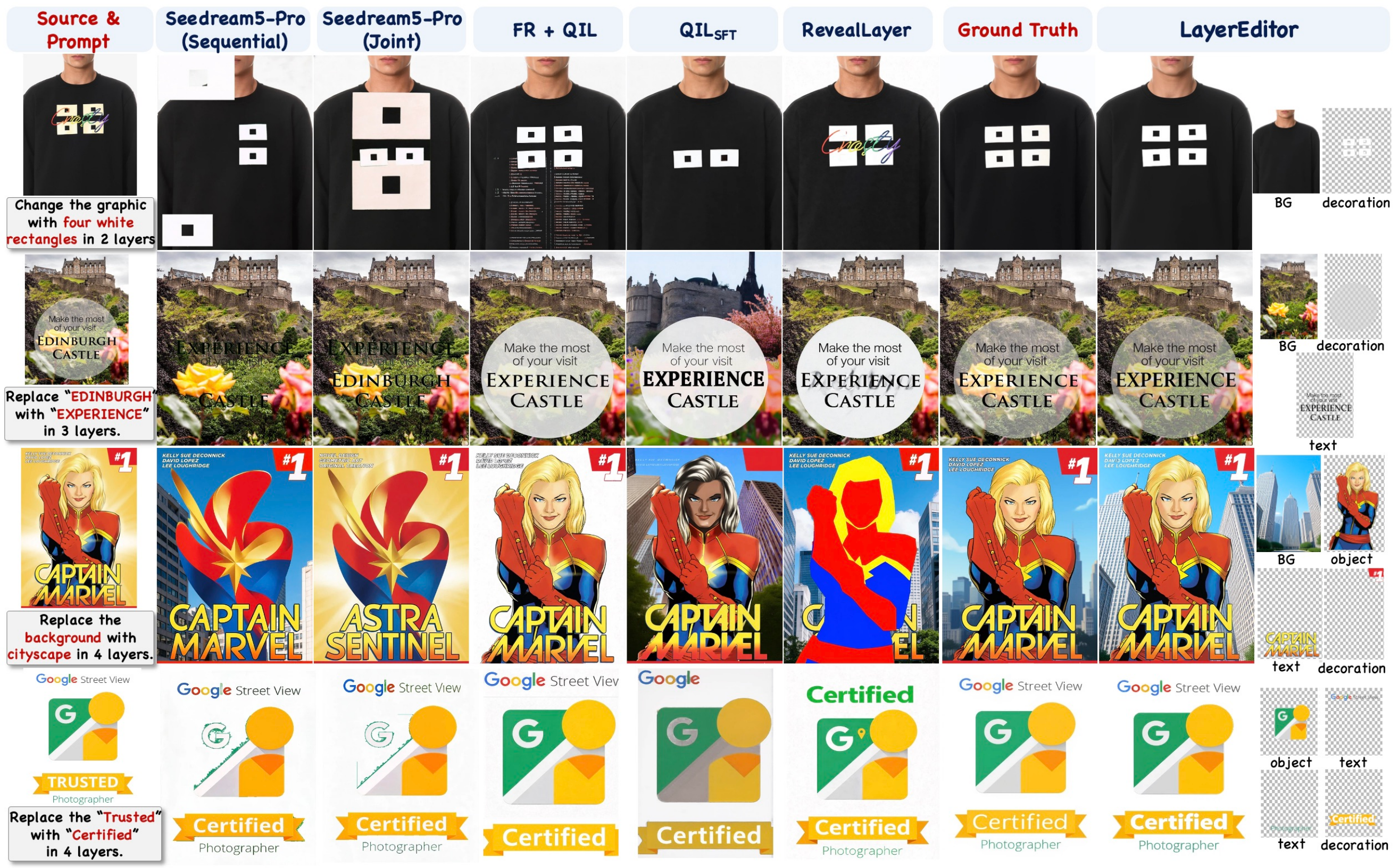}
\caption{Qualitative comparison of layered editing. All method outputs are rendered from the predicted RGBA stacks rather than the independently generated canvases. LayerEditor follows the editing instructions while producing coherent recompositions and reusable layers. }
\label{fig:overall_qualitative}
\end{figure}

\subsection{Editing Fidelity and Layer Reusability}
\label{sec:overall}

A key requirement of layered editing is to preserve the requested edit after recomposition while producing accurate individual layers. All FR-based cascades share the same edited canvases and therefore have identical Direct metrics. Table~\ref{tab:main} shows that decomposition fine-tuning improves Composite SSIM from 0.5584 to 0.6343 on \mbox{Crello-Testset} and from 0.6451 to 0.6927 on \mbox{ScaleEdit-Testset}. LayerEditor further achieves the best Composite SSIM/FID and Alpha IoU on both benchmarks. Compared with Seedream5-Pro (Joint), it improves Crello Composite SSIM by 5.7\% and reduces Composite FID by 38.8\%. On \mbox{ScaleEdit-Testset}, Composite SSIM increases from 0.8365 to 0.8839 and Alpha IoU from 0.6583 to 0.6802. These results support stronger recomposition fidelity and opacity agreement under the evaluated settings. We further compare Direct and Composite outputs in terms of their respective similarities to the edited target. On \mbox{ScaleEdit-Testset}, Seedream5-Pro (Seq.) drops from 0.8933 Direct SSIM to 0.7901 Composite SSIM, corresponding to an 11.6\% reduction. For LayerEditor, target SSIM decreases from 0.8925 to 0.8839, a relative reduction of 1.0\%, indicating a smaller loss of target similarity in the layered output.

Rubric-based evaluation provides complementary evidence of layer usability. LayerEditor achieves a mean score of 6.00 across the five dimensions on \mbox{Crello-Testset}, compared with 4.50 for Seedream5-Pro (Joint), the highest baseline mean. On \mbox{Crello-Testset}, its layer-segmentation score exceeds Seedream5-Pro (Joint) by 1.61 points, and its layer-composition score exceeds FR + QIL-Decomp-FT by 1.29 points. Figure~\ref{fig:overall_qualitative} further shows that competing methods often exhibit missing or duplicated content, mixed components, and transparency errors, whereas LayerEditor better preserves the requested edit while producing well-separated reusable RGBA layers.

\begin{figure}[tbp]
\centering
\includegraphics[width=\linewidth]{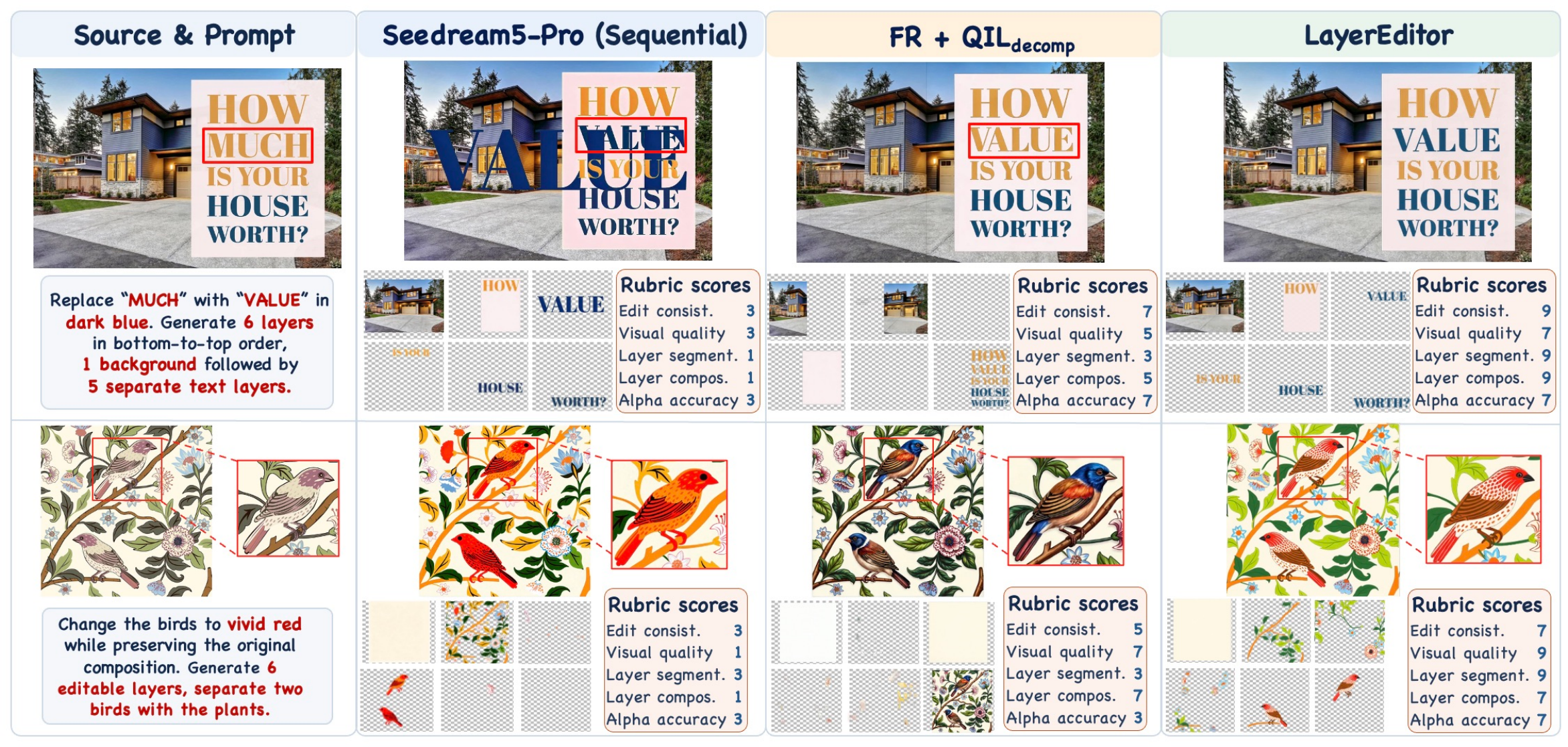}
\caption{Prompt-conditioned layer-content control. Both examples request six RGBA layers with specified layer content. LayerEditor performs the requested edits while assigning text rows and birds to the corresponding reusable layers.}
\label{fig:text_results}
\end{figure}

\subsection{Layer-Structure and Content Controllability}
\label{sec:reuse}
\label{sec:control}
To evaluate layer-count controllability, we construct a controlled setting from the 3,994 cases in \mbox{Crello-Testset} and \mbox{ScaleEdit-Testset}. For each case, the source image and edited appearance are fixed, while the reference layers are regrouped into different organizations of $2$--$10$ layers, with the requested count $N$ and layer-content descriptions updated accordingly. This setup isolates layer granularity and tests whether the model can represent the same edited content with different requested layer structures while preserving editing and layer quality. As shown in Table~\ref{tab:transfer}, LayerEditor achieves the highest Alpha IoU and rubric-based layer scores, exceeding Seedream5-Pro (Joint) by 0.0493 in Alpha IoU and 0.97 points in semantic grouping. The metrics are first averaged over the variants of each source document and then across documents.

\noindent\begin{minipage}{\linewidth}
\begin{minipage}[c]{.37\linewidth}
\normalsize\raggedright
Beyond controlling layer granularity, LayerEditor supports explicit control over the content assigned to individual layers. Figure~\ref{fig:text_results} presents two editing examples in which the prompt specifies both the visual modification and the desired layer organization. LayerEditor performs the requested text replacement and bird recoloring while separating individual text rows and birds into the specified layers. These results show that LayerEditor can control not only how finely an edited image is decomposed, but also which visual content is represented by each reusable layer.
\end{minipage}\hfill
\begin{minipage}[c]{.60\linewidth}
\makeatletter\def\@captype{table}\makeatother
\centering
\caption{Layer-structure controllability across different requested layer counts. Sem., Comp., and Alpha denote semantic grouping, composition, and Alpha scores. Time is observed latency per sample, without controlled hardware.}
\label{tab:transfer}
\begingroup\fontsize{8.5}{10}\selectfont\setlength{\tabcolsep}{1.4pt}\renewcommand{\arraystretch}{1.04}
\begin{adjustbox}{max width=\linewidth}
\begin{tabular}{@{}l*{6}{c}@{}}\toprule
&\multicolumn{2}{c}{\textcolor{accentblue}{Layer metrics}}&\multicolumn{3}{c}{\textcolor{accentpurple}{Rubric scores}}&\\
\cmidrule(lr){2-3}\cmidrule(l){4-6}
Method&\makecell{RGBA\\L1$\downarrow$}&\makecell{Alpha\\IoU$\uparrow$}&Sem.$\uparrow$&Comp.$\uparrow$&Alpha$\uparrow$&\makecell{Time$\downarrow$\\(s/sample)}\\\midrule
Seedream5-Pro (Seq.)&0.1184&0.4984&5.52&4.68&2.77&322.6\\
Seedream5-Pro (Joint)&\cellcolor{secondblue}\underline{0.1028}&\cellcolor{secondblue}\underline{0.5703}&\cellcolor{secondblue}\underline{5.91}&5.15&\cellcolor{secondblue}\underline{4.28}&250.0\\
FR + QIL&0.1625&0.2378&4.30&3.66&2.02&104.2\\
FR + QIL-Control&0.1175&0.4285&5.47&4.56&2.56&80.0\\
FR + QIL-Decomp-FT&0.1987&0.2554&5.65&\cellcolor{secondblue}\underline{5.47}&1.36&104.2\\
QIL-Edit-SFT&0.1148&0.3096&5.10&4.57&1.60&204.1\\
RevealLayer&0.4096&0.2208&3.21&3.75&1.11&\cellcolor{secondblue}\underline{43.7}\\
\midrule
\rowcolor{bestgreen}LayerEditor&\cellcolor{bestgreen}\textbf{0.0460}&\cellcolor{bestgreen}\textbf{0.6196}&\cellcolor{bestgreen}\textbf{6.88}&\cellcolor{bestgreen}\textbf{6.28}&\cellcolor{bestgreen}\textbf{6.98}&\textbf{41.5}\\\bottomrule
\end{tabular}\end{adjustbox}\endgroup
\end{minipage}
\end{minipage}\par\smallskip

\subsection{Layer Fidelity and Decomposition Capability}

This section evaluates LayerEditor on pure image-to-layer decomposition, where the input image is preserved and only its reusable RGBA layers are recovered. We construct the benchmark from the original Crello test split by regrouping compatible native layers and retaining 1,171 held-out images with $2$--$10$ reference layers. LayerEditor receives only the input image, a preservation instruction, and the requested layer count, without I2L-specific fine-tuning or reference layer descriptions. We compare against QIL, QIL-Decomp-FT, and the iterative LayerD baseline~\citep{suzuki2025layerd}, with evaluation details provided in Appendix~\ref{app:metrics}. LayerD follows its official inference procedure, with repeated decomposition to obtain the requested number of layers.

\begin{wrapfigure}{r}{.65\linewidth}
\centering
\includegraphics[width=\linewidth]{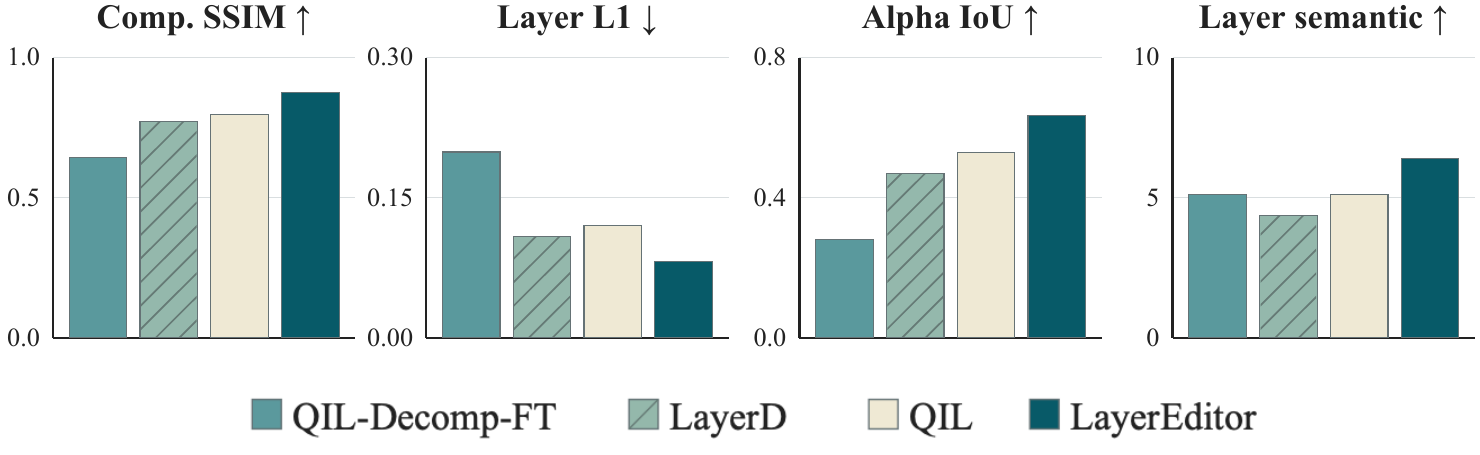}
\caption{Image-to-layer decomposition on Crello test set.}
    \label{fig:i2l}
\end{wrapfigure}
\begingroup\normalsize\raggedright
As shown in Figure~\ref{fig:i2l}, LayerEditor achieves a Composite SSIM of 0.8761, compared with 0.7963 for QIL. Its Alpha IoU of 0.6343 and layer-semantic score of 6.39 also exceed the corresponding QIL and QIL-Decomp-FT results. These observations support task transfer to image decomposition without I2L-specific fine-tuning. Figure~\ref{fig:text_results} provides examples of text integrity, object boundaries, and content isolation under explicit layer specifications.
\par\endgroup

\subsection{Ablation Study}
\label{sec:ablation}

\begin{table}[!ht]
\makeatletter\def\@captype{table}\makeatother
\centering
\begin{minipage}[c]{.42\linewidth}

\fontsize{8.3}{9.5}\selectfont
\setlength{\tabcolsep}{1.8pt}
\renewcommand{\arraystretch}{1.04}
\begin{tabularx}{\linewidth}{@{}Ycccc@{}}
\toprule
Model & \makecell{Edit\\Cons.$\uparrow$} & \makecell{Visual\\Qual.$\uparrow$} & \makecell{Layer\\Seg.$\uparrow$} & \makecell{RGBA\\L1$\downarrow$} \\
\midrule
\rowcolor{bestgreen}
LayerEditor         & \textbf{6.14} & \textbf{6.35} & \textbf{6.83} & \textbf{0.0350} \\
\midrule
- w/o C-EPD           & 4.67 & 5.12 & 5.01 & 0.0482 \\
- w/o Direct OPD      & 5.01 & 4.19 & 4.83 & 0.0513 \\
- w/o Comp. OPD       & 4.98 & 4.33 & 4.97 & 0.0492 \\
- w/o Alpha Dec.      & 5.12 & 5.30 & 5.67 & 0.0578 \\
\midrule
Base Model            & 3.93 & 4.13 & 3.77 & 0.0735 \\
\bottomrule
\end{tabularx}
\caption{Ablation study on \mbox{Crello-Testset}.}
\label{tab:ablation}
\end{minipage}
\hfill
\begin{minipage}[c]{.57\linewidth}

\centering
\includegraphics[width=.96\linewidth]{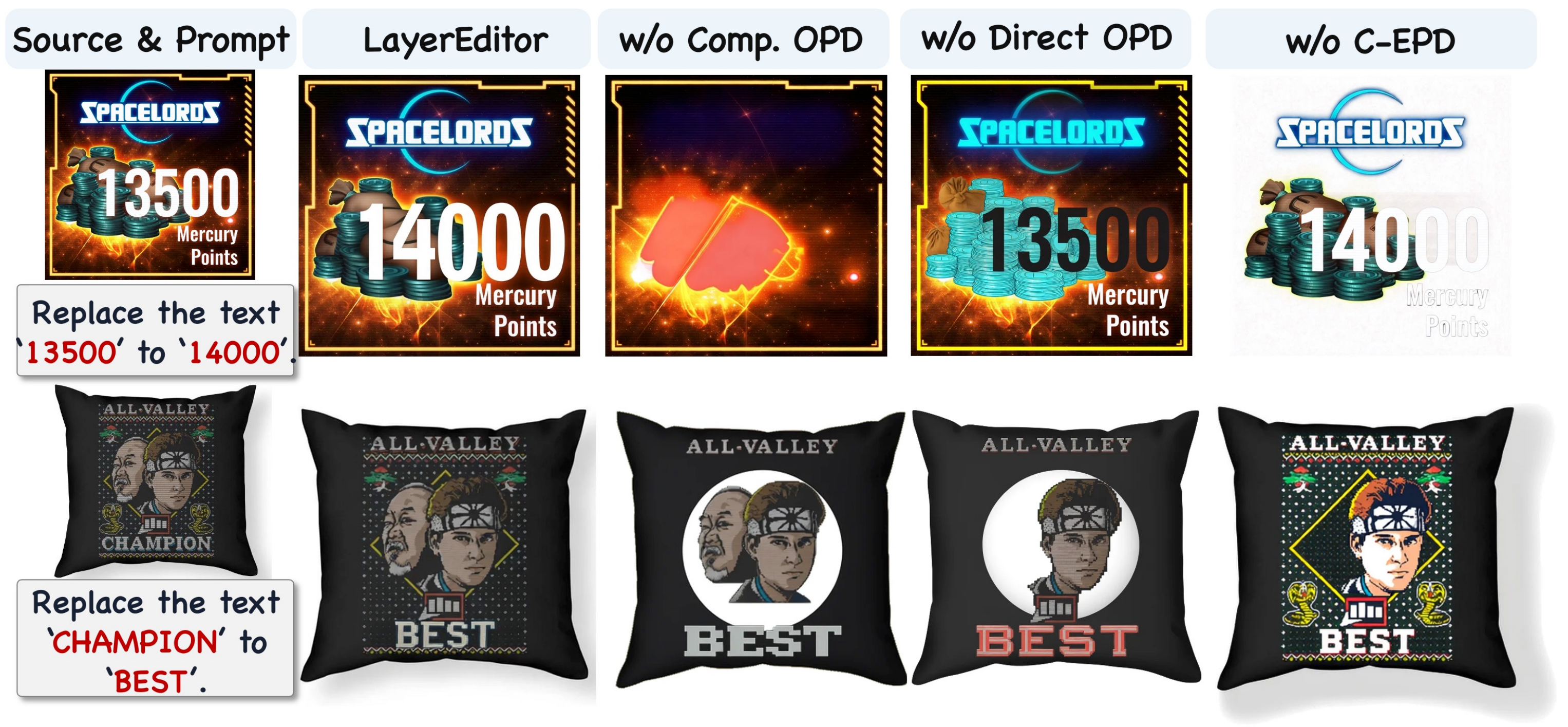}
\makeatletter
\def\@captype{figure}
\makeatother
\setlength{\abovecaptionskip}{2pt}
\setlength{\belowcaptionskip}{3pt}
\caption{Effect of C-EPD on recomposed outputs.}
\label{fig:ablation_qualitative}
\end{minipage}
\end{table}

We conduct ablation studies on \mbox{Crello-Testset} under matched backbone training steps and identical base configurations. The Base Model combines FireRed-1.1, In-Context RGB--Alpha Token Modeling, and flow matching, recovering Alpha by averaging the default pretrained VAE decoder's three output channels. LayerEditor further introduces the Alpha decoder together with the two components of C-EPD: Direct velocity guidance and pixel-space guidance on the decoded layer composition. As shown in Table~\ref{tab:ablation}, removing C-EPD reduces edit consistency from 6.14 to 4.67 and layer segmentation from 6.83 to 5.01. Compared with w/o C-EPD, either guidance term alone improves edit consistency but lowers visual-quality and layer-segmentation scores and increases RGBA L1. Combining both terms improves all four metrics, supporting their joint use for editing fidelity and layer quality. These ablations support the effectiveness of C-EPD under the evaluated training setup.Figure~\ref{fig:ablation_qualitative} further shows improved text editing and better preservation of surrounding content when both terms are used. The dedicated Alpha-decoding pathway improves layer reconstruction compared with averaging the outputs of the pretrained RGB decoder, reducing RGBA L1 from 0.0578 to 0.0350 and increasing layer segmentation from 5.67 to 6.83.

\section{Conclusions}
\label{sec:conclusion}
In this work, we introduce LayerEditor, a unified model that couples instruction-driven image editing with controllable generation of reusable RGBA layers. In-Context RGB--Alpha Token Modeling captures cross-layer dependencies and appearance--transparency interactions, while a dedicated Alpha decoder refines opacity for fine structures. C-EPD transfers the pretrained editing prior to both the edited canvas and composed layer predictions, helping preserve editing fidelity during layered adaptation. We further construct LayerEditor-100K for paired training supervision and establish dedicated benchmarks for evaluating layered image editing. Extensive experiments demonstrate strong recomposition fidelity, accurate layer separation, and flexible control over layer counts and content. By generating independently editable layers alongside the edited canvas, LayerEditor enables fine-grained revisions and content reuse beyond the initial edit.

\bibliography{references}
\bibliographystyle{layereditor}
\appendix
\setcounter{topnumber}{2}
\setcounter{totalnumber}{3}
\makeatletter
\setlength{\@fptop}{0pt}
\makeatother

\clearpage
\section*{Supplementary Material}
This supplementary material provides implementation details, dataset construction procedures, evaluation protocols, and additional experimental results. LayerEditor takes a source image, a requested layer count, and a textual editing instruction that may additionally specify the desired content of individual layers. These textual conditions are used only as model inputs, while RGBA layers, Alpha masks, and related annotations serve exclusively as training supervision or evaluation references.

\section{Implementation Details}
\label{app:implementation}
\subsection{Training and Inference Configuration}
LayerEditor adapts FireRed-Image-Edit-1.1 with rank-256 LoRA on the DiT backbone. Image and text encoders and the pretrained RGB decoder remain frozen. The Alpha decoder is trained separately before C-EPD and then frozen, as detailed in Appendix~\ref{app:latent_decoder}. Backbone adaptation begins with two epochs of flow matching, followed by joint flow-matching and C-EPD optimization. At inference, one denoising trajectory conditioned on $(I^{\mathrm{src}},u,N)$ produces the edited canvas and ordered RGB--Alpha fields, with no teacher calls. Images preserve their aspect ratio within a $1{,}048{,}576$-pixel budget. Table~\ref{tab:training_recipe} summarizes the backbone configuration, while Table~\ref{tab:decoder_recipe} gives the separate Alpha-decoder training settings.
\begin{table}[!ht]
\centering\small
\caption{Backbone training and inference hyperparameters.}
\label{tab:training_recipe}
\begin{tabular*}{\linewidth}{@{\extracolsep{\fill}}lclc@{}}
\toprule
\textbf{Parameter} & \textbf{Value} & \textbf{Parameter} & \textbf{Value} \\
\midrule
LoRA rank & 256 & Maximum pixel budget & $1024^2$ \\
Flow-matching warm-up & 2 epochs & Post-training budget & 3 epochs \\
Optimizer & AdamW & Learning rate & $2\times10^{-5}$ \\
LR schedule & Cosine & Minimum learning rate & $5\times10^{-6}$ \\
LR warm-up & 200 updates & Maximum C-EPD weight & 0.1 \\
Main-model batch size & 64 & C-EPD interval & 5 updates \\
Inference steps & 30 & CFG scale & 4.0 \\
\bottomrule
\end{tabular*}
\end{table}

\paragraph{Benchmark conditions and optional prompt enhancement.}
\label{app:prompt_enhancer}
Each benchmark case specifies a source image, editing instruction, requested count $N$, and ordered layer-content descriptions. QIL, QIL-Decomp-FT, and RevealLayer use count-based decomposition without benchmark content descriptions. QIL-Control, QIL-Edit-SFT, Seedream5-Pro, and LayerEditor receive the same requested count and ordered layer-content specifications. These form count-based and content-conditioned comparison groups. Reference target pixels and Alpha masks are never generation inputs. Training descriptions are aligned with the annotated target stack. Benchmark layer-content descriptions are constructed by our layer-annotation pipeline and undergo manual screening before evaluation. The content-conditioned benchmarks therefore evaluate generation under a supplied layer plan. I2L transfer uses the source image, a preservation request, and the requested count, without reference layer descriptions.

For in-the-wild requests, an optional Prompt Enhancer helps users formulate layer descriptions and normalize the request into \texttt{Instruction}, \texttt{Preserve}, ordered layer descriptions, and \texttt{Output} fields. It is an input-authoring aid rather than an additional source of benchmark conditions.

\subsection{Alpha Decoder}
\label{app:latent_decoder}
\label{app:decoder_losses}
The Alpha decoder retains the VAE topology with a single-channel Alpha-logit head, initialized with zero weights and bias $-3.5$. Alpha maps are repeated across three channels, scaled to $[-1,1]$, and encoded by the frozen VAE. Decoding uses only Alpha latents, without RGB features or bounding-box masks. The w/o Alpha Decoder and Base Model variants instead decode each Alpha latent with the unchanged pretrained RGB VAE decoder and average its three output channels to obtain a single Alpha map, without dedicated opacity training. The dedicated Alpha decoder recovers opacity as:
\begin{equation}
A_i=\sigma\!\left(g_\phi(z_i^A)\right),
\label{eq:alpha_decode}
\end{equation}
\begin{samepage}
where $\sigma$ is the sigmoid function. Layer reconstruction $\mathcal L_{\mathrm{rec}}$ combines regional Alpha errors with support, boundary, area, multiscale, and black/white-rendering terms. The full objective adds composition, perceptual, and stage-specific adversarial supervision:
\begin{equation}
\mathcal L_\phi=\mathcal L_{\mathrm{rec}}+\lambda_{\mathrm{comp}}\mathcal L_{\mathrm{comp}}+\lambda_{\mathrm{perc}}\mathcal L_{\mathrm{perc}}+\lambda_{\mathrm{adv}}\mathcal L_{\mathrm{adv}}.
\label{eq:decoder_objective}
\end{equation}
\end{samepage}
Here, $\mathcal L_{\mathrm{comp}}$ measures document-level L1 reconstruction of premultiplied color and accumulated opacity. $\mathcal L_{\mathrm{perc}}$ uses per-layer black/white-background LPIPS~\citep{zhang2018lpips} with AlexNet v0.1 at up to $262{,}144$ pixels. Stages 2--3 use $\lambda_{\mathrm{comp}}=1$ and $\lambda_{\mathrm{perc}}=0.05$. The PatchGAN term uses $\lambda_{\mathrm{adv}}=0.005$ from step 4,000 of stage 2, with discriminator weight 0.10. These auxiliary terms are disabled outside the specified stages. Layer and composition losses are normalized by the global counts of sampled layers and documents, respectively.

Training proceeds through single-layer reconstruction, full-document supervision, and latent-noise adaptation (Table~\ref{tab:decoder_recipe}). Stage 1 updates only the output head for its first 500 steps. Stage 3 disables adversarial training and perturbs RGB/Alpha latents with Gaussian noise scaled by their per-layer standard deviations ($\sigma=0.01$). All stages use bfloat16, with the RGB VAE frozen. Alpha-decoder training is completed before C-EPD. Both decoders remain frozen during backbone adaptation, while retaining input gradients for compositional guidance. Final RGB and Alpha outputs are rendered in back-to-front order:
\begin{equation}
\mathcal C_{-1}=B,\qquad \mathcal C_i=A_i\odot R_i+(1-A_i)\odot\mathcal C_{i-1},\qquad \mathcal C=\mathcal C_{N-1}.
\label{eq:composition}
\end{equation}
Here, $B$ is the background canvas, set to white for evaluation. The Direct canvas is not part of the layer stack.

\begin{table}[!ht]
\centering\small
\caption{Alpha-decoder configuration. Values separated by slashes follow stage order.}
\label{tab:decoder_recipe}
\setlength{\tabcolsep}{3.5pt}
\begin{adjustbox}{max width=\linewidth}
\begin{tabular*}{\linewidth}{@{\extracolsep{\fill}}lclc@{}}
\toprule
\textbf{Parameter} & \textbf{Value} & \textbf{Parameter} & \textbf{Value} \\
\midrule
Maximum pixel budget & $1024^2$ & Spatial alignment & 16 pixels \\
Batch per GPU & 1 document & Effective batch size & 32 documents \\
GPU count & 8 & Gradient accumulation & 4 steps \\
Stage duration & 3 / 5 / 2 epochs & Initial learning rate & $10^{-5}$ / $5\times10^{-6}$ / $2\times10^{-6}$\\
Encoding microbatch & 2 layers & Decoding microbatch & 32 layers \\
Composition weight & 1.0 in stages 2--3 & Perceptual weight & 0.05 in stages 2--3 \\
Latent noise scale & 0.01 in stage 3 & Adversarial onset & Stage 2, step 4,000 \\
\bottomrule
\end{tabular*}
\end{adjustbox}
\end{table}

\subsection{C-EPD Training}
\label{app:opd}
The on-policy rollout uses 10 nodes sampled from the 30-step inference sigma grid, with two nodes drawn from each of five temporal segments. Each active C-EPD update samples two queries from high-, medium-, and low-noise bands with probabilities 0.2, 0.5, and 0.3. The rollout states are detached. At each selected noisy state, the student recomputes velocities for the Direct, RGB, and Alpha fields with gradients. The frozen teacher predicts its image-level velocity from the same canvas state, source image, instruction, and noise level, without gradients.

At each query, we decode the local clean RGB--Alpha estimates and compose them on white. The teacher's Direct prediction yields a detached, decoded target. Equation~\ref{eq:composite_opd} combines Direct velocity and composite pixel errors, averaged over the two queries. Student-side decoding remains differentiable, and the composite is not re-encoded. Appendix~\ref{app:pixel_composition} details the computation. Flow matching remains active on all fields at every update.

The first two epochs use flow matching only. The local counter $k$ then counts post-training optimizer updates, with C-EPD enabled every five updates. Its weight is $\lambda_{\mathrm C}(k)=0.1\min(k/500,1)$, and $\lambda_{\mathcal S}=0.05$ weights the composite pixel term. This ramp is separate from the 200-update learning-rate warm-up in Table~\ref{tab:training_recipe}. On active updates, the weighted C-EPD loss is added to flow matching as in Equation~\ref{eq:lora_objective}. Other updates use flow matching alone. The teacher remains frozen and is used only during training. Algorithm~\ref{alg:layereditor_training} summarizes the procedure. Flow matching averages the squared velocity error over all elements of the joint sequence, as in Equation~\ref{eq:flow}.

\begin{figure}[tbp]
\makeatletter\def\@captype{algorithm}\makeatother
\centering
\caption{LayerEditor training. The pretrained editing model is used as the distillation teacher.}
\label{alg:layereditor_training}
\smallskip
\footnotesize
\textbf{Input:} training set $\mathcal D_{\mathrm{train}}$, frozen editor $\theta_0$, LoRA parameters $\theta$, pretrained Alpha decoder $\phi$, total epochs $E$, and warm-up epochs $E_{\mathrm{warm}}=2$. The local counter $k$ starts at zero after warm-up. \\[4pt]
\textbf{Fields:} $\Omega=\{\mathcal D,R_0,A_0,\ldots,R_{N-1},A_{N-1}\}$. Both decoders are frozen throughout backbone training.
\medskip\hrule\medskip
\setlength{\tabcolsep}{3pt}\renewcommand{\arraystretch}{1.08}
\begin{tabularx}{\linewidth}{@{}r X@{}}
1 & \textbf{for} $e=1,\ldots,E$ \textbf{do}\\
2 & \quad\textbf{for} $(I^{\mathrm{src}},u,N,Z^*)\sim\mathcal D_{\mathrm{train}}$ \textbf{do}\\
3 & \qquad Sample $t\sim p(t)$ and $\epsilon\sim\mathcal N(0,I)$.\\
4 & \qquad Construct $Z_t=(1-\sigma_t)Z^*+\sigma_t\epsilon$ and $v_t^*=\epsilon-Z^*$.\\
5 & \qquad Predict $\{v_\theta^X\}_{X\in\Omega}$ from the joint noisy state and $(I^{\mathrm{src}},u,N)$.\\
6 & \qquad Compute the joint flow-matching loss using Equation~\ref{eq:flow}.\\
7 & \qquad $\mathcal L_\theta\gets\mathcal L_{\mathrm{flow}}$. If $e>E_{\mathrm{warm}}$, set $k\gets k+1$.\\
8 & \qquad\textbf{if} $e>E_{\mathrm{warm}}$ and the local post-training update $k$ is divisible by 5 \textbf{then}\\
9 & \qquad\quad Sample two queries $q\sim\mathcal Q$ from the detached 10-node student rollout.\\
10 & \qquad\quad Recompute student velocities from the complete joint query state with gradients.\\
11 & \qquad\quad Query the frozen teacher on $\bar z_q^{\mathcal D}$ with the source image, editing instruction, and $\sigma_q$.\\
12 & \qquad\quad Estimate $\hat z_{0|q}^X\gets\bar z_q^X-\sigma_qv_{q,S}^X$ for every layer field $X\in\{R_i,A_i\}$.\\
13 & \qquad\quad Decode $\hat R_i\gets\mathcal D_R(\hat z_{0|q}^{R_i})$ and $\hat A_i\gets\mathcal D_A(\hat z_{0|q}^{A_i})$, retaining input gradients.\\
14 & \qquad\quad Compose $\widehat{\mathcal C}_S$ on white using Equation~\ref{eq:pixel_composite}, without re-encoding.\\
15 & \qquad\quad $\mathcal I_T\gets\operatorname{sg}[\mathcal D_R(\bar z_q^{\mathcal D}-\sigma_qv_{q,T}^{\mathcal D})]$.\\
16 & \qquad\quad Average the two query losses in Equation~\ref{eq:composite_opd} to obtain $\mathcal L_{\mathrm{C\text{-}EPD}}$.\\
17 & \qquad\quad $\mathcal L_\theta\gets\mathcal L_{\mathrm{flow}}+\lambda_{\mathrm C}(k)\mathcal L_{\mathrm{C\text{-}EPD}}$.\\
18 & \qquad\textbf{end if}\\
19 & \qquad Update $\theta$ using $\nabla_\theta\mathcal L_\theta$.\\
20 & \quad\textbf{end for}\\
21 & \textbf{end for}\\
22 & \textbf{return} $\theta,\phi$.\\

\end{tabularx}
\medskip\hrule
\end{figure}

\needspace{8\baselineskip}
\subsection{\texorpdfstring{Local Decoding and Pixel-Space Composition}{Local Decoding and Pixel-Space Composition}}
\label{app:pixel_composition}

Here ``local'' refers to a selected state on the denoising trajectory. We decode full-canvas predictions at that state, rather than spatial crops. Under the linear flow parameterization~\citep{lipman2023flowmatching}, $z_\sigma=(1-\sigma)z_0+\sigma\epsilon$ and $v^*=\epsilon-z_0$ imply $z_0=z_\sigma-\sigma v^*$. At detached student query states, predicted velocities yield approximate clean layer latents:
{
\begin{equation}
\hat z_{0|q}^{R_i}=\bar z_q^{R_i}-\sigma_qv_{q,S}^{R_i},\qquad \hat z_{0|q}^{A_i}=\bar z_q^{A_i}-\sigma_qv_{q,S}^{A_i}.
\label{eq:local_clean_layers}
\end{equation}
}
The frozen RGB and Alpha decoders recover $\hat R_i$ and $\hat A_i$ from these estimates. RGB values are expressed in $[0,1]$, and the Alpha decoder's sigmoid produces continuous opacity in $[0,1]$. We use the same output scaling, white background, and original back-to-front order as at inference. Equation~\ref{eq:pixel_composite} then performs source-over composition, with the visible contribution of layer $i$ weighted by $w_i=\hat A_i\prod_{j=i+1}^{N-1}(1-\hat A_j)$. The uncovered background contributes $B\prod_{j=0}^{N-1}(1-\hat A_j)$. These weights act on decoded RGB, not on latent velocities. The Direct canvas is not a layer in this recurrence.

The frozen teacher is queried on the corresponding Direct state. Its local clean estimate is decoded using the same frozen RGB decoder to form the target:
{
\begin{equation}
\mathcal I_T=\operatorname{sg}\!\left[\mathcal D_R(\bar z_q^{\mathcal D}-\sigma_qv_{q,T}^{\mathcal D})\right],\qquad \mathcal L_{\mathcal C}^{\mathrm{pixel}}=\frac{1}{d_{\mathcal C}}\|\widehat{\mathcal C}_S-\mathcal I_T\|_2^2.
\label{eq:pixel_teacher_target}
\end{equation}
}
Here $d_{\mathcal C}=3HW$ for an $H\times W$ RGB composite. The teacher target is a local clean prediction, not a completed denoising sample. The pixel loss supervises the rendered composition without re-encoding it or assuming that the noisy layer states equal the Direct state.

Rollout construction and teacher-target computation are detached. At each selected query, the student is recomputed with gradients, and its local clean estimates pass through frozen decoders and differentiable composition. Freezing decoder parameters does not stop gradients with respect to their inputs. Thus, the pixel loss can guide both RGB and Alpha predictions through their visible effects on the composite. Fully occluded content remains supervised by the layer-wise flow-matching objective. Decoding is needed only at the selected queries, not at every rollout step.

\section{LayerEditor-100K Construction}
\label{app:data}
\subsection{Data Construction Pipeline}
\label{app:data_routes}
\begin{figure}[tbp]
    \centering
    \includegraphics[width=\linewidth]{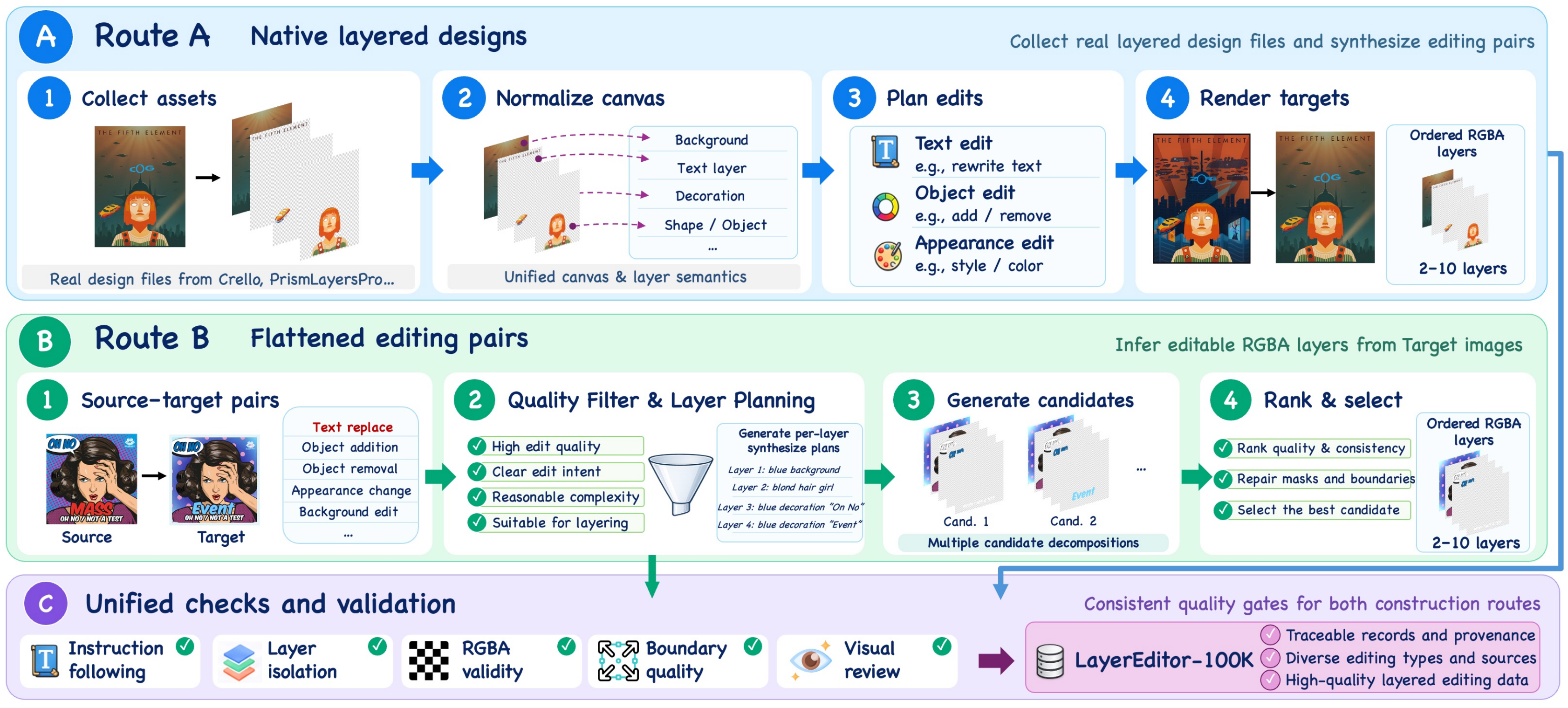}
    \caption{LayerEditor-100K construction pipeline.
    Two complementary routes create editing pairs with reusable RGBA annotations from native designs and flattened editing pairs. Shared validation checks editing fidelity, layer quality, and reconstruction before export.}
    \label{fig:data_pipeline}
\end{figure}

LayerEditor-100K combines two routes for obtaining paired editing supervision and reusable target layers (Figure~\ref{fig:data_pipeline}). Every accepted record contains a source image, instruction, edited target, and ordered full-canvas RGBA stack. Layer annotations provide RGBA supervision and aligned semantic descriptions for training the structured prompt interface (Appendix~\ref{app:prompt_enhancer}).

\paragraph{Route A: native layered designs.}
Crello and PrismLayersPro provide native assets with positions, stacking order, and transparency~\citep{yamaguchi2021canvasvae,chen2025prismlayers}. We first render and consolidate the assets into $2$--$10$ ordered layers, then plan the edit while keeping key text, subjects, and target regions separate where possible. Layout changes, movement, and deletion are handled deterministically, while generative editing is used for text redrawing, object replacement, and appearance modification. The edited content is finally combined with preserved RGBA layers to form the target stack, followed by semantic and visual quality verification.

\paragraph{Route B: flattened editing pairs.}
\label{app:scaleedit}
ScaleEdit source--target pairs~\citep{chen2026scaleedit} are screened for effective edits, readable content, and suitability for layered editing. A planner proposes layer roles and granularity, and QIL~\citep{yin2025qwenlayered} and Seedream5-Pro~\citep{bytedance2026seedream5pro} generate decomposition candidates. Candidates are standardized to full-canvas RGBA stacks with a complete bottom background and transparent foreground support. Repairable failures receive at most two rounds of local correction for text defects, missing content, leakage, or Alpha artifacts, followed by full-stack review. Severe failures are rejected, and repairs are retained only when they improve the candidate. The original target is preserved when reconstruction passes. A repaired composition becomes the target only after independent verification of instruction execution and content preservation. Both versions and the revision reason are retained.

\paragraph{Benchmark construction.}
\mbox{Crello-Testset} and \mbox{ScaleEdit-Testset} are constructed separately from the native-design and flattened-pair source groups, with no training-source overlap. Their main editing results are evaluated separately. To construct the two controllability settings, we combine all samples from both benchmarks, modify the editing prompt or requested layer count, and revise each target and reference stack to match the modified request. Count-only changes regroup compatible layers while preserving the intended edited appearance. Instruction changes update both edited content and its reference layers. Each derived case fixes its source, conditions, target, and reference stack before inference. Target pixels and reference RGBA arrays are withheld from generation. Their annotations provide the supplied layer-content specifications.

\subsection{Layer Annotation and Quality Control}
\label{app:pe_template}
\label{app:data_validation}
Native layer metadata is updated after editing in Route A, while a planner infers target-layer structure from the edited pair in Route B. Each layer is annotated with its index, role, content, edit status, and bounding box. Semantic roles and content descriptions populate the training prompt, while geometric metadata supports offline validation. Figure~\ref{fig:pe_template} summarizes the four acceptance conditions. Quality control follows a common sequence of structure check, semantic check, layer QA, reconstruction, and deduplication. Structure checks validate count, dimensions, and order. Semantic checks compare source, instruction, and target, including transcription for text edits. Layer QA inspects isolated RGBA components and Alpha boundaries, and reconstruction checks compare the exact composition with the intended target. 

\begin{figure}[tbp]
\centering
\fbox{\begin{minipage}{0.93\linewidth}
\small\raggedright
\textbf{Offline target-layer annotation and acceptance.}\par\smallskip
\textbf{Count and order:} exactly $N$ nonempty layers in valid back-to-front order.\par\smallskip
\textbf{Content isolation:} complete intended components without duplication, cross-layer leakage, or unrelated background residue.\par\smallskip
\textbf{Valid Alpha:} transparent foreground support with appropriate continuous opacity at fine structures and semitransparent boundaries.\par\smallskip
\textbf{Accurate recomposition:} the stack reconstructs the target while retaining the requested edit and unchanged content.\par\smallskip
\textbf{Record:} layer index, role, visible content, edit status, and bounding box. Semantic descriptions support structured textual control. RGBA references and geometric metadata support supervision and validation.
\end{minipage}}
\caption{Offline layer annotation and quality criteria. Shared checks validate the target stack produced by either construction route.}
\label{fig:pe_template}
\end{figure}

\subsection{Dataset Statistics and Split}
LayerEditor-100K contains 100,000 accepted editing pairs from Crello, PrismLayersPro, and ScaleEdit. The dataset includes 48,329 text edits, 36,858 object edits, and 14,813 poster-level appearance edits, with $2$--$10$ target layers per example. Counts refer to unique accepted records, excluding rejected candidates and repeated training exposures. Training and test documents do not overlap. All edits and layer organizations derived from the same source document remain in the same split. The original Crello documents used for I2L evaluation are also excluded from training-data construction.

\mbox{Crello-Testset} contains 2,994 documents from native Crello and PrismLayersPro designs. \mbox{ScaleEdit-Testset} contains 1,000 documents from flattened ScaleEdit pairs with recovered layers. Layer-count control and instruction control are each evaluated on the combined full pool of 3,994 documents. In the layer-count setting, each source document typically yields two to four organizations at compact, intermediate, or detailed granularities. Each organization has a corresponding count, ordered content description, and reference stack, with the source image and edited target preserved. The 3,994 count denotes source documents, not the number of derived requests. Scores are first averaged over the evaluated variants of each source document and then over source documents, so documents with more variants receive no additional weight. Instruction-control requests instead update the edited target and reference stack to match the new instruction. Layer-count results are reported in Table~\ref{tab:transfer}, and instruction-control results in Table~\ref{tab:instruction_control}. Every derived example remains in the test split of its source document.

\section{Evaluation Details}
\label{app:evaluation}
\subsection{Baselines}
\label{app:baselines}
Except for Seedream5-Pro, which is accessed through its hosted API, all baseline results are obtained from our own runs of the corresponding models. Pretrained or fine-tuned configurations are specified below. LayerD results in Figure~\ref{fig:i2l} are also reproduced by us rather than taken from published tables. The FR editor in cascaded baselines is pretrained, whereas LayerEditor receives layered-editing adaptation. These are system-level comparisons and do not isolate joint modeling from editing-data adaptation. Count-only baselines additionally differ in available conditions. Comparisons with QIL-Control, QIL-Edit-SFT, and Seedream5-Pro hold the supplied layer-content specifications fixed.
\paragraph{Seedream5-Pro (Seq.).}
We access Seedream5-Pro through its API~\citep{bytedance2026seedream5pro}. Seq. uses two calls: the first produces the edited image, and the second decomposes it using the requested count and layer-content descriptions. No task-specific fine-tuning is applied.

\paragraph{Seedream5-Pro (Joint).}
Joint uses one API call with the source image, editing instruction, requested count, and the same layer-content descriptions supplied to LayerEditor. It requests editing and layer generation together. The returned stack supplies Composite outputs. No independent Direct canvas is exported in this setting.

\paragraph{FR + QIL.}
FireRed edits the source, and pretrained QIL decomposes the result using the requested layer count~\citep{firered2026rededit,yin2025qwenlayered}. The FR stage uses 40 sampling steps, guidance of 4, and aspect-ratio-preserving dimensions aligned to multiples of 16 within a $1024^2$ pixel budget. All FR-based cascades reuse these edited canvases within each setting, including the supplementary instruction-control experiment.

\paragraph{FR + QIL-Control.}
QIL-Control replaces QIL in the decomposition stage~\citep{diffsynth2026layeredcontrol}. It uses text-conditioned extraction with the same ordered layer-content descriptions supplied to LayerEditor, and the requested count specifies the size of the assembled stack. No task-specific fine-tuning is applied.

\paragraph{QIL-Decomp-FT.}
QIL is fine-tuned on LayerEditor-100K targets and RGBA annotations following the official decomposition-training pipeline. It decomposes FR edits for layered editing or the source image for I2L, using the requested count without benchmark layer-content descriptions.

\paragraph{QIL-Edit-SFT.}
This adapted QIL baseline learns canvas and layer generation directly from source images, editing instructions, requested counts, and layer-content descriptions. It uses the same content specifications as LayerEditor, with paired targets and RGBA stacks as supervision.

\paragraph{RevealLayer.}
RevealLayer~\citep{wang2026reveallayer} decomposes the FR edit without task-specific fine-tuning. Its bounding-box guidance uses the benchmark reference annotations. The requested count guides repeated extraction and stack assembly, without benchmark layer-content descriptions. Post-generation correspondence supports evaluation when the recovered organization differs from the reference.

\paragraph{LayerD.}
LayerD~\citep{suzuki2025layerd} is included in the I2L comparison. We follow the official inference procedure and repeatedly invoke decomposition to assemble a stack with the requested target layer count. These sequential calls incur substantial time cost as more layers are extracted. They are part of the decomposition pipeline, not a single joint prediction of all layers. We use the pretrained model without task-specific fine-tuning.

\subsection{Metrics and Layer Matching}
\label{app:metrics}
\textbf{Rendering and image metrics.} Direct metrics compare the exported canvas $\mathcal D$ with the target. Composite metrics compare the target with $\mathcal C$, recomputed from all exported layers in generated order on white. SSIM uses target-aligned images with the same size. FID compares Inception-v3 feature distributions over the evaluated set and is interpreted only within a benchmark, not across benchmark sizes. Rendering uses the final exported stack without evaluation-time registration or reference-guided repair. Correspondence matching for layer metrics does not reorder the rendered stack. Differences between Direct and Composite SSIM compare their respective similarities to the target. They do not measure SSIM between $\mathcal D$ and $\mathcal C$.

\textbf{Layer correspondence and RGBA L1.}
We request the specified layer count from every model. We first use Hungarian assignment to establish one-to-one correspondences between nonempty predicted and reference layers, maximizing the total IoU of Alpha supports thresholded at 0.05. An empty support union has similarity zero. No additional acceptance threshold is applied, so low-overlap assigned pairs still contribute their measured reconstruction errors. Only identically zero-Alpha output slots are excluded from this assignment, and small nonempty layers are retained.

Unassigned predicted and reference layers, including empty output slots, are then paired in their respective original stacking orders. If the residual counts differ, zero-RGBA placeholders pad the shorter list. Every original layer is used exactly once. Let $\mathcal P$ contain all assigned, ordered residual, and padded pairs. For four-channel arrays in $[0,1]$, we compute:
{
\begin{equation}
\operatorname{RGBA\ L1}=\frac{1}{N_{\mathrm{ref}}}\sum_{(P,G)\in\mathcal P}\frac{\|P-G\|_1}{4HW}.
\label{eq:rgba_l1_complete}
\end{equation}
}
Here $N_{\mathrm{ref}}$ is the reference layer count. Each pair contributes its actual four-channel L1, including remaining layers compared with empty placeholders. There is no manually assigned error or failure penalty. For equal-count stacks, the metric is the mean per-layer error. A fixed reference-count denominator prevents extra zero-RGBA slots from diluting the score. If extra nonzero layers are generated, their errors also enter the sum. The metric uses stored RGB and Alpha directly, without premultiplication. It measures reference-array agreement and can differ for visually equivalent layers whose fully transparent pixels store different RGB values. We therefore interpret it alongside opacity and Composite metrics.

Continuous Alpha IoU is $\sum_p\min(A_p,A_p^*)/\sum_p\max(A_p,A_p^*)$ on the same pairs, with zero for an empty union and the same reference-count normalization. Per-sample image and layer scores are averaged within each source document before aggregation across documents. FID remains a distribution-level statistic over the evaluated image set. Matching does not reorder the stack used to render $\mathcal C$. Rubric evidence retains original layer indices and includes residual content and empty slots when assessing order, content assignment, and completeness.

As a separate diagnostic, we count layers with Alpha support below 0.1\% of the canvas as empty or near-empty. Such layers constitute \textbf{1.02\%} of all predicted layers for LayerEditor and \textbf{10.37\%} for Seedream5-Pro (Seq.) across the two pooled benchmarks. This area threshold is used only for the diagnostic statistic, not to discard small layers from the reconstruction metric.

\paragraph{I2L and baseline scope.}
The I2L target is the input image itself. LayerEditor, QIL, and QIL-Decomp-FT are compared on the same 1,171 held-out documents. LayerD obtains its stack through repeated extraction, with correspondence established after generation. The requested count determines the iterative extraction target, while the content of the recovered layers is determined by the decomposition procedure. Figure~\ref{fig:i2l} reports layer reconstruction separately from image reconstruction. All plotted results, including LayerD, come from our own evaluation runs. LayerD's recovered granularity and post-generation correspondence differ from the count-controlled comparisons. We therefore base the transfer conclusion on the shared QIL protocol rather than treating the LayerD layer scores as a directly controlled ranking.

\paragraph{Retries, incomplete outputs, and timing.}
Seedream5-Pro uses at most five API calls per evaluation sample, where a sample is one editing-and-layering plan with a fixed prompt and requested layer count. The cap includes remedial calls for missing or invalid outputs and applies separately to each plan, rather than being shared across all variants of a source document. The nominal Seq. and Joint protocols require two and one calls, respectively. Five documents from \mbox{Crello-Testset} remain without usable Seedream5-Pro outputs because of API refusals, including copyright-related restrictions. Its Crello metrics are computed on the remaining 2,989 documents, using the corresponding references. Missing cases are excluded rather than assigned artificial scores. Other methods are evaluated on the full benchmark with the same rendering and metric definitions. Thus, the Seedream5-Pro Crello scores use an available-case subset. Table~\ref{tab:transfer} reports the mean observed time to complete editing and layer generation for one evaluation sample, in seconds per sample (s/sample). Timing is averaged per plan, whereas quality scores are aggregated within source documents before averaging across documents. Hardware, service overhead, and repeated-call costs are not controlled, so these times are not a matched model-speed ranking.

\paragraph{Ablation protocol.}
All variants use RGB--Alpha Token Modeling with identical backbone optimization steps and matched base configurations. All variants use the same joint flow-matching objective in Equation~\ref{eq:flow}. The dedicated Alpha decoder has its own reconstruction-training schedule, reported separately in Appendix~\ref{app:latent_decoder}. LayerEditor includes the dedicated Alpha decoder, Direct velocity supervision, and composite pixel supervision against the decoded teacher target. The w/o C-EPD variant keeps the Alpha decoder but removes both teacher constraints, giving a flow-matching-only control with the same decoding architecture. The w/o Direct OPD variant retains only composite pixel guidance, while w/o Comp.\ OPD retains only Direct velocity matching. Both still generate the Direct canvas and the full layer stack under flow matching. The w/o Alpha Decoder variant retains both constraints but uses the unchanged pretrained three-channel VAE decoder without dedicated opacity training. Base Model uses the same default decoder and flow matching alone, without C-EPD. Flow matching remains active in every variant. These controls evaluate C-EPD as a whole and the contribution of each teacher constraint when the other is retained. In w/o Alpha Decoder, the same default decoder and three-channel averaging are used for local compositional guidance and final layer rendering. Its comparison with the dedicated decoder therefore measures the combined effect of opacity reconstruction on supervision and output quality, particularly at boundaries and within color regions. It does not isolate decoder architecture from additional opacity training. These ablations also do not separate on-policy query selection from teacher supervision, as no matched off-policy control is included.

\subsection{Rubric-Based Evaluation}
\label{app:rubric}
\label{app:gemini}
A single Gemini-3.1-Pro evaluator assigns separate scores for edit consistency, visual quality, layer segmentation, layer composition, and Alpha accuracy under the fixed rubric in Figure~\ref{fig:rubric_instructions}. All reported benchmark scores use the recomputed Composite $\mathcal C$ as the edited-output evidence. The Direct canvas is excluded from rubric scoring. The source image and instruction establish the requested edit and preservation requirements, while the corresponding target, reference layers, and Alpha maps support dimension-specific judgments. All methods use the same score anchors from 1 to 10, with explicit anchors at 1, 3, 5, 7, and 10.

For each layer-count variant, evaluation uses the target and reference stack associated with that request. Layer assessment considers content isolation, the generated stacking order, and agreement with the requested content assignments, using the matching procedure in Appendix~\ref{app:metrics}. Empty layers and unmatched-layer placeholders are included in the evidence supplied to the VLM, which judges their effect on content completeness, layer organization, and editing fidelity. Composite-based assessment captures their effect on the complete output. The reference-free instructions in Figure~\ref{fig:rubric_instructions} are not used for the reported benchmark scores.

\begin{figure}[hbp]
\centering
\includegraphics[width=\linewidth]{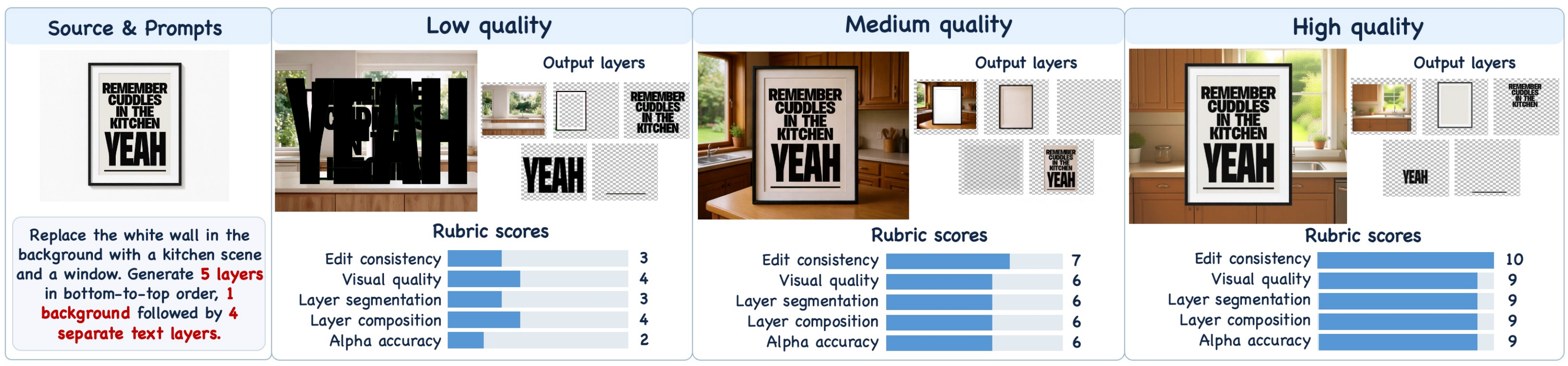}
\par\smallskip
\begin{minipage}{\linewidth}\small
\caption{Rubric score examples. For the same editing request, progressively improved outputs receive higher scores across editing fidelity, layer separation, composition, and Alpha quality. The examples illustrate the rubric's discriminative behavior rather than agreement with human ratings.}
\label{fig:rubric_examples}
\end{minipage}
\end{figure}

Figure~\ref{fig:rubric_examples} illustrates the score anchors with selected low-, medium-, and high-quality results. These examples show how visible defects are reflected in the scores, rather than validating agreement with human judgments. The model-based scores complement the reference-based image and layer metrics. The mean across five dimensions discussed in the main text is an unweighted descriptive summary, not an additional evaluator output. Semantic grouping in layer-count control and layer semantics in I2L are task-specific applications of the layer-segmentation criterion. They assess whether the extracted components form coherent semantic units at the requested granularity, using the same score anchors.

\begin{figure}[p]
\centering
\begingroup
\setlength{\fboxsep}{9pt}
\fcolorbox{accentblue}{white}{%
\begin{minipage}{0.93\linewidth}
\footnotesize\raggedright
{\color{accentblue}\bfseries Rubric Instructions for Layered Editing Evaluation}\par\smallskip
\hrule\smallskip
\textbf{Role and objective.} Evaluate instruction-driven editing and reusable RGBA layers. Score each assigned dimension independently from 1 to 10. Use the source as the preservation reference and the target as the intended edit. Do not combine dimensions into a weighted overall score.\par\smallskip
\textbf{Evidence assigned to each request.}
\begin{itemize}\setlength{\itemsep}{1pt}\setlength{\topsep}{1pt}
\item \textbf{Joint editing.} Inspect the source, instruction, target, and recomputed composition $\mathcal C$. Use only $\mathcal C$ as the edited-output evidence. Do not use the Direct canvas.
\item \textbf{Layers.} Inspect predicted and reference RGBA sheets, $\mathcal C$, and the target for the requested layer organization. Use the supplied correspondences to assess content, and retain the original layer indices when judging order and assignment. Inspect empty layers and unmatched-layer placeholders alongside the complete stack, and account for missing or extra content in the corresponding scores.
\item \textbf{Alpha.} Compare predicted and reference opacity sheets without substituting RGB appearance for mask accuracy.
\end{itemize}
\textbf{Score anchors.} A score of 1 denotes unusable or contradictory output, 3 denotes major failure, 5 denotes recognizable content with substantial defects, 7 denotes mostly correct output with localized defects, and 10 denotes no meaningful visible discrepancy. Use intermediate scores when evidence falls between these anchors.\par\smallskip
\textbf{Evaluation dimensions.}
\begin{itemize}\setlength{\itemsep}{1pt}\setlength{\topsep}{1pt}
\item \textbf{Edit consistency} checks the requested content, wording, attributes, and position while penalizing unintended changes.
\item \textbf{Visual quality} checks readability, layout, object integrity, detail, color blending, and visible artifacts.
\item \textbf{Layer segmentation} checks complete, unique, independently editable components and penalizes leakage, clipping, and duplication.
\item \textbf{Layer composition} checks whether the exported stack reconstructs the target without gaps, ghosting, displacement, or incorrect occlusion.
\item \textbf{Alpha accuracy} checks support, holes, leakage, thin boundaries, antialiasing, and semitransparency against reference masks.
\end{itemize}
\textbf{When references are unavailable.} This optional mode is not used for the reported benchmark scores. Without a target image, assess $\mathcal C$ against $I^{\mathrm{src}}$ and $u$ for instruction execution and source preservation. Without reference layers or masks, judge component completeness, isolation, and recomposition from the predicted stack. Do not infer unseen ground truth. Report reference-free judgments separately from reference-based scores.\par\smallskip
\textbf{Rendering and validity.} Recompute $\mathcal C$ from exported layers on white. Show RGBA sheets on checkerboards while retaining the original continuous opacity of Alpha sheets. Visible failures receive low scores. Visible generation failures, including empty or missing layers, must affect the scores. Invalid judgments are reserved for unavailable or corrupted evaluation evidence. API cases with no usable output follow the available-case protocol in Appendix~\ref{app:metrics}.\par\smallskip
\textbf{Output and task-specific scope.} Return a score and evidence-grounded rationale for each dimension. In the layer-count setting, assess grouping and reconstruction against the target and reference layers for each requested count. In I2L decomposition, assess source reconstruction and layer quality rather than editing-instruction execution.
\end{minipage}}
\endgroup
\caption{Instructions for VLM-based evaluation. Edit consistency, visual quality, layer segmentation, layer composition, and Alpha accuracy are scored separately using the evidence available for each setting.}
\label{fig:rubric_instructions}
\end{figure}

\section{Additional Results}
\subsection{Instruction Controllability}
\label{app:paired}

We evaluate instruction controllability on the combined 3,994 source documents from \mbox{Crello-Testset} and \mbox{ScaleEdit-Testset}. For each source image, we vary the editing instruction and construct the corresponding target image and reference RGBA stack, updating the layer count or organization when required by the new edit. Each prediction is evaluated only against the target and layer structure associated with its instruction. Scores are averaged within each source document before aggregation across the full source pool. This setting tests whether both the edited content and the generated layer organization respond consistently to changes in user intent. As shown in Table~\ref{tab:instruction_control}, LayerEditor achieves the highest Direct/Composite SSIM of 0.9018/0.8765 and Alpha IoU of 0.5823, together with the lowest RGBA L1 of 0.0344. The rubric-based results show the same trend. LayerEditor achieves an edit-consistency score of 5.54 compared with 3.32 for Seedream5-Pro (Joint), and a layer-segmentation score of 6.33 compared with 5.11 for FR + QIL-Control. It also maintains the highest visual-quality, layer-composition, and Alpha-accuracy scores, with a composition score of 5.68 versus 4.34 for FR + QIL-Decomp-FT. These results demonstrate that LayerEditor can adapt both the edited appearance and reusable layer structure to different instructions.

\begin{table}[!ht]
\centering
\caption{Instruction control on 3,994 documents from both benchmarks. Image, layer, and five rubric dimensions follow Table~\ref{tab:main}. FR-based cascades share the same Direct canvas.}
\label{tab:instruction_control}

\begingroup\fontsize{7.5}{8.8}\selectfont\setlength{\tabcolsep}{2.6pt}\renewcommand{\arraystretch}{1.12}
\begin{adjustbox}{max width=\linewidth}
\begin{tabular}{@{}l*{9}{c}@{}}\toprule
&\multicolumn{2}{c}{\textcolor{accentblue}{Image SSIM $\uparrow$}}&\multicolumn{2}{c}{\textcolor{accentteal}{Layer metrics}}&\multicolumn{5}{c}{\textcolor{accentpurple}{Rubric scores $\uparrow$ (1--10)}}\\
\cmidrule(lr){2-3}\cmidrule(lr){4-5}\cmidrule(l){6-10}
Method&Direct&Comp.&\makecell{RGBA\\L1$\downarrow$}&\makecell{Alpha\\IoU$\uparrow$}&\makecell{Edit\\consist.}&\makecell{Visual\\quality}&\makecell{Layer\\segment.}&\makecell{Layer\\compose.}&\makecell{Alpha\\accuracy}\\\midrule
Seedream5-Pro (Seq.)&\cellcolor{secondblue}\underline{0.8611}&0.7165&0.0675&0.4728&3.01&3.77&4.01&3.03&2.68\\
Seedream5-Pro (Joint)&--&\cellcolor{secondblue}\underline{0.7914}&\cellcolor{secondblue}\underline{0.0563}&\cellcolor{secondblue}\underline{0.5646}&\cellcolor{secondblue}\underline{3.32}&3.93&4.81&3.74&\cellcolor{secondblue}\underline{3.98}\\
FR + QIL&0.6113&0.4777&0.1435&0.1945&1.16&3.16&3.96&2.67&1.72\\
FR + QIL-Control&0.6113&0.5827&0.0984&0.3502&1.45&\cellcolor{secondblue}\underline{4.51}&\cellcolor{secondblue}\underline{5.11}&3.87&2.33\\
FR + QIL\textit{-Decomp-FT}&0.6113&0.5972&0.1733&0.2123&1.41&4.33&4.68&\cellcolor{secondblue}\underline{4.34}&1.14\\
QIL\textit{-Edit-SFT}&0.7294&0.7046&0.1978&0.2104&1.22&3.59&3.63&3.32&1.13\\
RevealLayer&0.6113&0.5578&0.2725&0.2531&1.19&3.00&2.64&2.21&1.09\\
\midrule
\rowcolor{bestgreen}LayerEditor&\cellcolor{bestgreen}\textbf{0.9018}&\cellcolor{bestgreen}\textbf{0.8765}&\cellcolor{bestgreen}\textbf{0.0344}&\cellcolor{bestgreen}\textbf{0.5823}&\cellcolor{bestgreen}\textbf{5.54}&\cellcolor{bestgreen}\textbf{5.74}&\cellcolor{bestgreen}\textbf{6.33}&\cellcolor{bestgreen}\textbf{5.68}&\cellcolor{bestgreen}\textbf{4.43}\\
\bottomrule\end{tabular}\end{adjustbox}\endgroup
\end{table}

\subsection{Additional Qualitative Results}
\label{app:qualitative}
\begin{figure}[tbp]
\centering
\includegraphics[width=\linewidth]{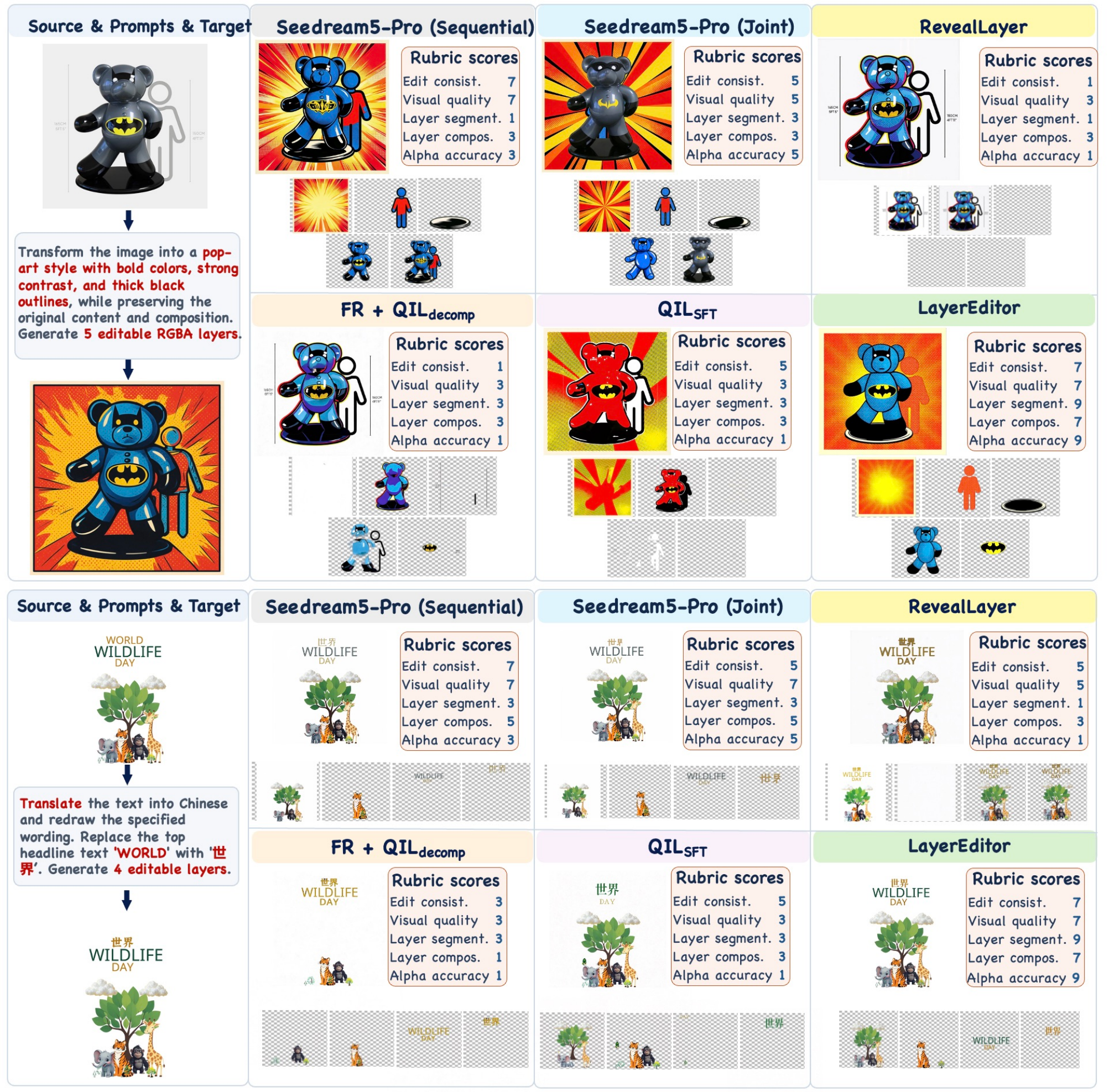}
\caption{Instruction-following editing and controllable layer decomposition on \mbox{Crello-Testset}. Prompts specify the edit and layer count. Large output images show recompositions, and checkerboard panels show individual RGBA layers.}
\label{fig:additional_qualitative}
\end{figure}

\begin{figure}[tbp]
\centering
\includegraphics[width=\linewidth]{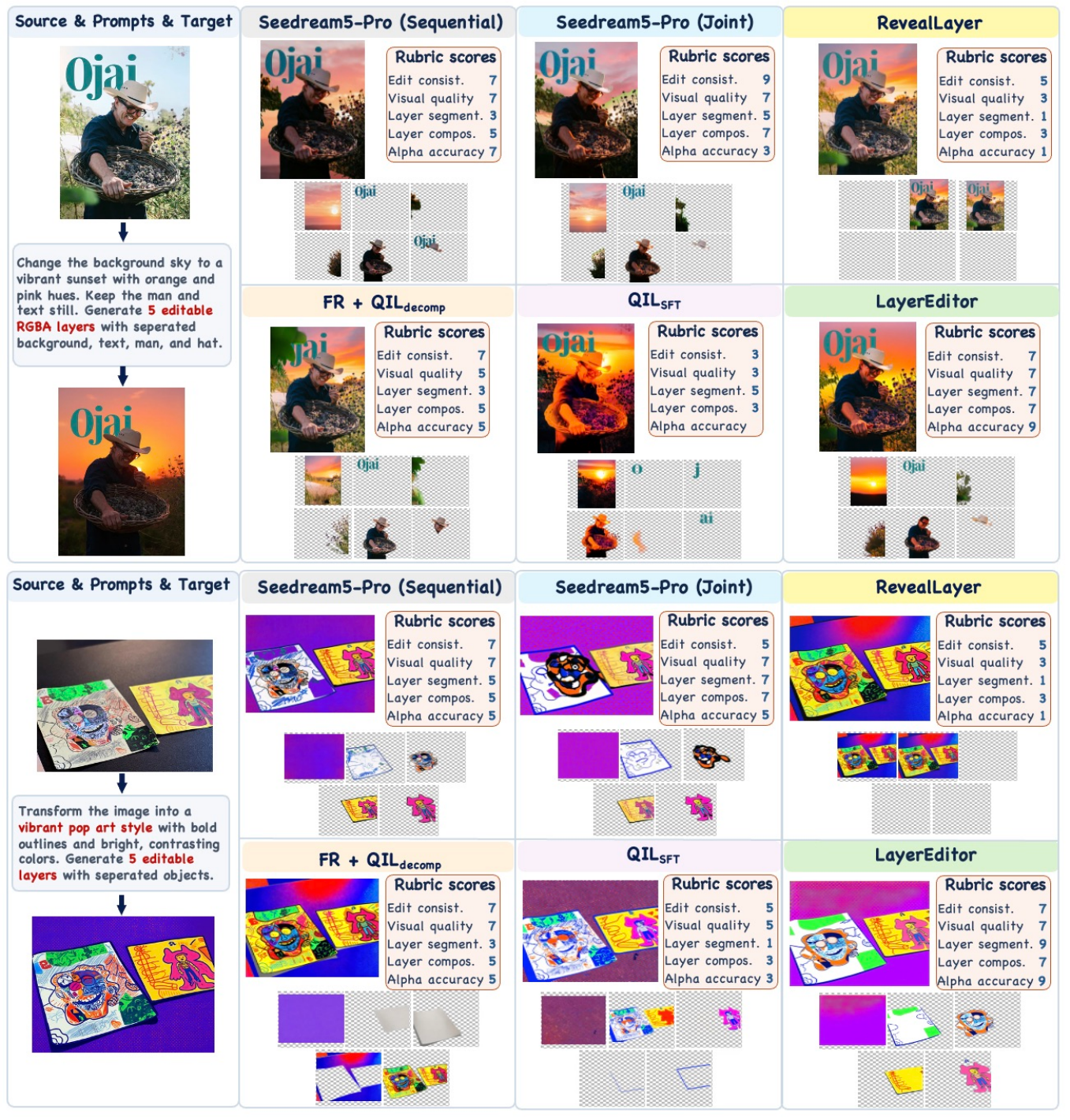}
\caption{Instruction-following editing and content-controlled layer decomposition on \mbox{ScaleEdit-Testset}. Prompts specify the edit, layer count, and desired content separation. Large output images show recompositions, and checkerboard panels show individual RGBA layers.}
\label{fig:layer_details}
\end{figure}

Figures~\ref{fig:additional_qualitative} and~\ref{fig:layer_details} illustrate instruction-following edits together with controllable layer decomposition on \mbox{Crello-Testset} and \mbox{ScaleEdit-Testset}. The examples cover pop-art restyling, text replacement, and background editing, with prompts specifying the layer count and, where applicable, the desired layer content. LayerEditor follows the requested appearance or text changes while separating text, objects, and backgrounds into reusable RGBA layers. In Figure~\ref{fig:additional_qualitative}, the edited text and scene components remain separately accessible. Figure~\ref{fig:layer_details} further illustrates the separation of the edited sky, text, person, and hat, as well as individual objects in a restyled scene. The recomposed outputs retain the requested edits, while the layer sheets reveal how the content is organized under the supplied specifications. Together, these examples illustrate editing fidelity and layer-level controllability within the same generation process.

\section{Limitations}
\label{app:limitations}
LayerEditor is trained and evaluated with at most ten RGBA layers, leaving larger stacks unexplored. Thin structures, complex transparency, and semitransparent boundaries can still produce artifacts during layer movement or recomposition. Text remains rasterized rather than represented as editable font or vector primitives, limiting downstream typography manipulation. The current results do not establish lossless source reconstruction or artifact-free downstream reuse.

Quantitative evaluation measures layer isolation, opacity, and recomposition in the evaluated layouts. We do not systematically test reconstruction after layer manipulation or validate VLM scores against human judgments. Editing-prior retention is evaluated on the constructed benchmarks rather than an independent general-editing test. Automatically constructed annotations and model-based assessments may contain errors and model-specific biases.

\section{Ethics Statement}
LayerEditor enables instruction-driven image editing and reusable layered representations. As with other image editing systems, it may be misused to alter visual content deceptively or modify copyrighted assets without authorization. Our work is intended for research on controllable image editing and structured visual representations. LayerEditor-100K is constructed from existing research datasets and design resources, and we preserve source provenance throughout data processing. We do not intentionally collect additional private user information. Automated vision-language models are used for semantic verification and quality assessment, and we combine model-based evaluations with deterministic checks and structured quality-control procedures. Because layered outputs make subsequent visual manipulation easier, responsible deployment should consider content provenance, copyright, and safeguards against deceptive editing.

\end{document}